\documentclass[10pt,journal,compsoc]{IEEEtran}
\ifCLASSOPTIONcompsoc
\else
\fi
\ifCLASSINFOpdf
\else
\fi

\usepackage{graphicx}
\usepackage{amsmath}
\usepackage{amssymb}
\usepackage{subfigure}
\usepackage{algorithm,algorithmic}
\usepackage{bm}
\usepackage{multirow}
\usepackage{cite}
\usepackage{hhline}
\usepackage{tabulary}
\usepackage{url}
\usepackage{epsfig}
\usepackage{epstopdf}
\usepackage{color}

\newcommand{\argmax}{\operatornamewithlimits{argmax}}
\newcommand{\argmin}{\operatornamewithlimits{argmin}}
\newcommand{\ssum}{\operatornamewithlimits{\sum\sum}}

\newcommand{\etal}{\emph{et~al.~}}
\newcommand{\ie}{\emph{i.e.~}}
\newcommand{\eg}{\emph{e.g.~}}
\def\mathbi#1{\textbf{\em #1}}

\begin{document}
%
\title{Person Re-identification by saliency Learning}
%
%
%
%

\author{Rui~Zhao, ~\IEEEmembership{Student Member,~IEEE,~}
        Wanli~Oyang, ~\IEEEmembership{Member,~IEEE,~} and~\\
        Xiaogang~Wang,~\IEEEmembership{Member,~IEEE}
\IEEEcompsocitemizethanks{\IEEEcompsocthanksitem R. Zhao, W. Ouyang, and X. Wang are with the Department of Electronic Engineering, the Chinese University of Hong Kong, Hong Kong.\protect\\
E-mail: \{rzhao, wlouyang, xgwang\}@ee.cuhk.edu.hk
}
\thanks{}}

%
%

\markboth{Manuscript draft}%
{Shell \MakeLowercase{\textit{et al.}}: Bare Demo of IEEEtran.cls for Computer Society Journals}
%



\IEEEtitleabstractindextext{%
\begin{abstract}
Human eyes can recognize person identities based on small salient regions, i.e. human saliency is distinctive and reliable  in pedestrian matching across disjoint camera views. However, such valuable  information is often hidden when computing similarities of pedestrian images with existing approaches. Inspired by our user study result of human perception on human saliency, we propose a novel perspective for person re-identification based on learning human saliency and matching saliency distribution. The proposed saliency learning and matching framework consists of four steps: (1) To handle misalignment caused by drastic viewpoint change and pose variations, we apply adjacency constrained patch matching to build dense correspondence between image pairs. (2) We propose two alternative methods, i.e. K-Nearest Neighbors and One-class SVM, to estimate a saliency score for each image patch, through which distinctive features stand out without using identity labels in the training procedure. (3) saliency matching is proposed based on patch matching. Matching patches with inconsistent saliency brings penalty, and images of the same identity are recognized by minimizing the saliency matching cost. (4) Furthermore, saliency matching is tightly integrated with patch matching in a unified structural RankSVM learning framework. The effectiveness of our approach is validated on the VIPeR dataset and the CUHK01 dataset. Our approach outperforms the state-of-the-art person re-identification methods on both datasets. 

\end{abstract}

}

\maketitle

\IEEEdisplaynontitleabstractindextext

%
\IEEEpeerreviewmaketitle

\section{Introduction}
Person re-identification \cite{gong2013person, vezzani2013people, bedagkar2014survey} is to match pedestrians observed from non-overlapping camera views based on image appearance. It has important applications in video surveillance such as human retrieval, human tracking, and activity analysis. It saves a lot of human efforts on exhaustively searching for a person from large amounts of images and videos. Nevertheless, person re-identification is a very challenging task. A person observed in different camera views undergoes significant variations on viewpoints, poses, and illumination, which make intra-personal variations even larger than inter-personal variations. Image blurring, background clutters and occlusions also cause additional difficulties. 

\begin{figure}[!t]
\centering
\includegraphics[width = 0.49\textwidth]{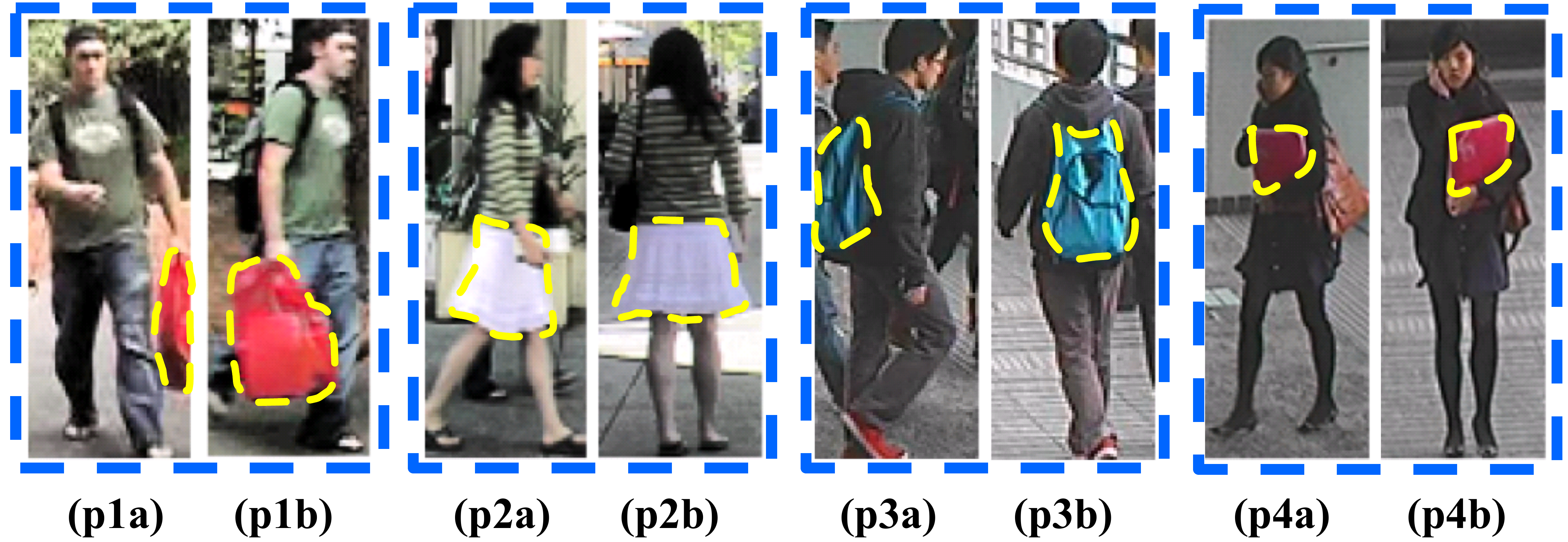}
\caption{\small{Salient region could be a body part or a carrying accessory. Some salient regions of pedestrians are highlighted with yellow dashed boundaries.} }
    \label{fig:intro1}
\end{figure}

Variations of viewpoints and poses commonly exist in person re-identification, and cause misalignment between images. In Figure \ref{fig:intro1}, the lower right region of $(p1a)$ is a red bag, while a leg appears in this region in $(p1b)$; the central region of $(p3a)$ is an arm, while it becomes a backpack in $(p3b)$.  
Most existing methods \cite{prosser2010person, zheng2011person, li2012human, dikmen2011pedestrian, mignon2012pcca} match pedestrian images by first computing the difference of feature vectors and then the similarities based on such difference vectors, which is problematic due to the spatial misalignment. In our work, patch matching is employed to handle misalignment, and it is integrated with saliency matching to improve the discriminative power and robustness to spatial variation.

Salient regions in pedestrian images provide valuable information in identification. However, if they are small in size, saliency information is often overwhelmed by other features when computing similarities of images. In this paper, \emph{saliency} means regions with attributes that 1) make a person \emph{distinctive} from their candidates, and 2) are \emph{reliable} in finding the same person across camera views. In many cases, humans can easily recognize matched pedestrian pairs because they have distinct features. For example, in Figure \ref{fig:intro1}, person $p1$ takes a red bag, $p2$ dresses bright white skirt, $p3$ takes a blue bag, and $p4$ carries a red folder in arm. These features are discriminative in distinguishing one person from others. Intuitively, if a body part is salient in one camera view, it usually remains salient in another camera view. Therefore, saliency also has view invariance.

Salient regions are not limited to body parts (such as clothes and trousers), but also include accessories (such as baggages, folders and umbrellas as shown in Figure \ref{fig:intro1}), which are often considered as outliers and removed in existing approaches. 
Our computation of saliency is based on the comparison with images from a large scale reference dataset rather than a small group of persons. Therefore, it is quite stable in most circumstances.

\begin{figure}[!t]
\centering
\includegraphics[width = 0.49\textwidth]{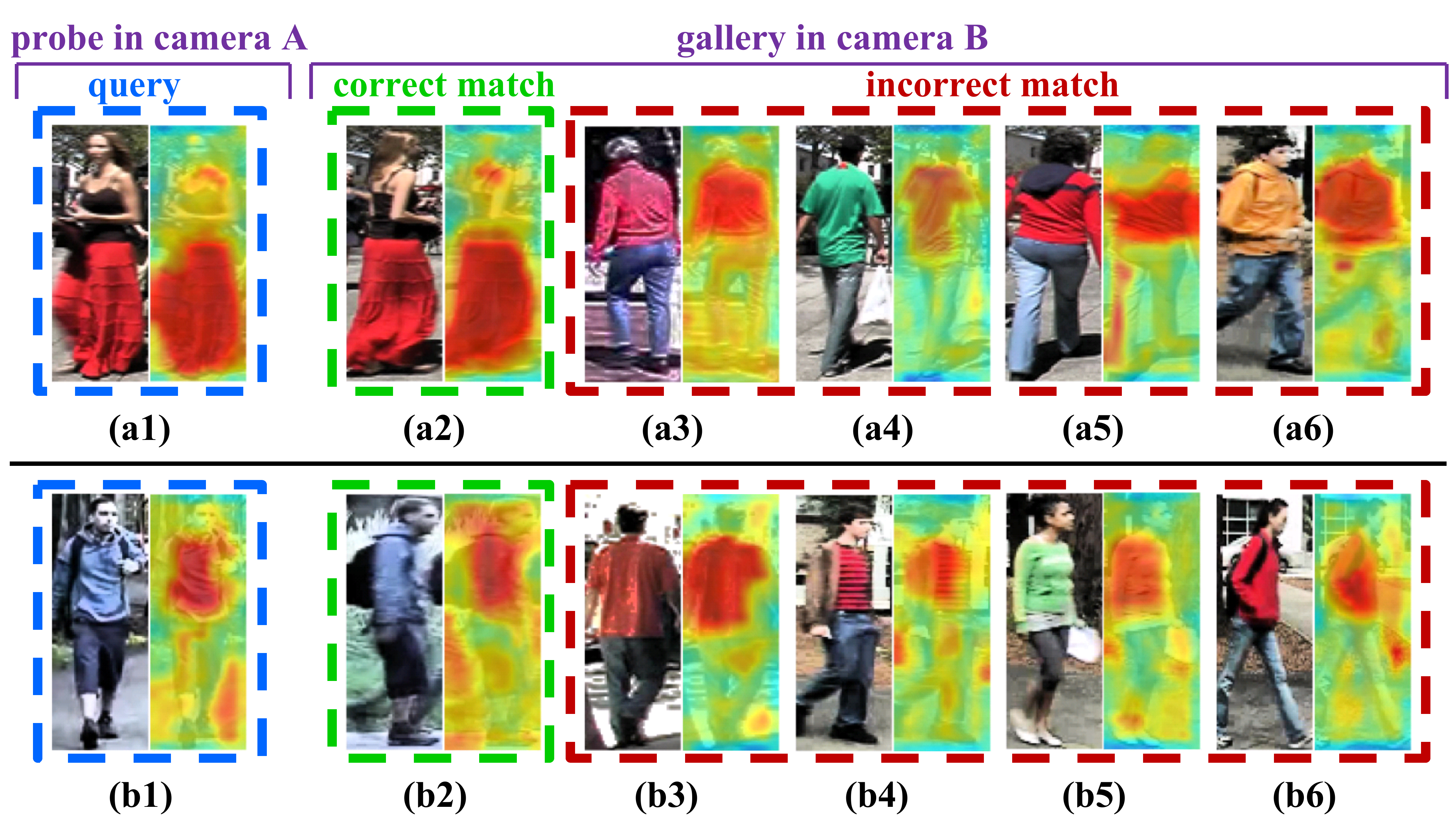}
\caption{\small{Illustration of saliency matching with examples. saliency map of each pedestrian image is shown. \textbf{Best viewed in color}.}}
    \label{fig:intro2}
\end{figure}

We observe that images of the same person captured from different camera views have some invariance property on their spatial distributions of saliency, like pair $(a1, a2)$ in Figure \ref{fig:intro2}. 
Since the person in image $(a1)$ shows saliency in her dress while others $(a3)$-$(a6)$ are salient in blouses, they can be well distinguished simply from the spatial distributions of saliency. 
Therefore, not only the visual features from salient regions are discriminative, the spatial distributions of human saliency also provide useful information in person re-identification. 
Such information can be encoded into patch matching. If two patches from two images of the same person are matched, they are expected to have similar saliency values; otherwise such matching brings penalty on saliency matching. 
In the second row in Figure \ref{fig:intro2}, the query image $(b1)$ shows a similar saliency distribution as those of gallery images. 
In this case, visual similarity needs to be considered. This motivates us to relate saliency matching penalty to the visual similarity of two matched patches. 

\section{Our approach}
Although saliency plays an important role in person re-identification, it has not been well explored in literature. In this paper, a novel framework of human saliency learning and matching is proposed for person re-identification. Our major contributions can be summarized from the following aspects.

\begin{enumerate}
  \item We propose a way of estimating saliency based on human perception through user study. It is estimated from the number of trials that a human subject recognizes a query image from a candidate pool only based on a local region. It shows that most pedestrian images can be matched by humans from local salient regions without looking at whole images. 
	The saliency estimated from user study is compared with the result of our saliency computation model. Compared with general image saliency detection methods \cite{borji2012exploiting, goferman2012context}, our proposed saliency computation has much stronger correlation with human perception in person re-identification.  

  \item A computation model is proposed to estimate the probabilistic saliency map. Different from general image saliency detection methods, it is specially designed for person re-identification, and has the following properties. 1) It is robust to changes of viewpoints, poses and articulation. 2) Distinct patches are considered as salient only when they are matched and distinct in both camera views. 3) Human saliency itself is a useful descriptor for pedestrian matching. For example, a person only with salient upper body and a person only with salient lower body must be different identities.



 \item We formulate person re-identification as a saliency matching problem. Dense correspondences between patches are established by patch matching based on visual similarity, and matching patches with inconsistent saliency brings cost. Images of the same person are recognized by minimizing the saliency matching cost, which depends on both locations and visual similarity of matched patches. 

  \item saliency matching and patch matching are tightly integrated into a unified structural RankSVM  framework. Structural RankSVM has good training efficiency given a  large number of rank constraints in person re-identification. Our approach transforms the original high-dimensional visual feature space to a 80 times lower dimensional saliency feature space to further improve training efficiency and also avoid overfitting. 

\end{enumerate}



\section{Related Works} \label{sec:relatedworks}

Existing works on person re-identification mainly focus on two aspects: 1) \emph{features and representations}, and 2) \emph{distance metric}. A review can be found in \cite{gong2013person}. 

\subsection{Features and Representations}

A lot of research efforts \cite{wang2007shape,farenzena2010person,bak2010person,cheng2011custom, cheng2014person,ma2012bicov,malocal2012fisher, ma2014discriminative,zheng2009associating,zheng2014group,xu2013human} have been devoted to exploiting discriminative features in person re-identification. Wang \etal \cite{wang2007shape} proposed shape and appearance context to model the spatial distributions of appearance relative to body parts in order to extract discriminative features robust to misalignment. Farenzena \emph{et al}. \cite{farenzena2010person} proposed the Symmetry-Driven Accumulation of Local Features (SDALF) by exploiting the symmetry property in pedestrian images to handle view variation. Bak \emph{et al}. \cite{bak2010person}, Xu \etal \cite{xu2013human} and Cheng \etal \cite{cheng2011custom, cheng2014person} applied human part models and pictorial structures to cope with pose variations by establishing the spatial correspondence. 
Ma \emph{et al}. \cite{ma2012bicov,malocal2012fisher, ma2014discriminative} developed the BiCov descriptor based on the Gabor filters and the covariance descriptor to handle illumination change and background variation. Zheng \etal \cite{zheng2009associating,zheng2014group} used the contextual visual cues from surrounding people to enrich human signatures. 
Information on salient regions exploited in our work can be integrated with many of these feature designs by putting more weights on features from salient regions.

Features vary in their usefulness in person matching, and some works have been done on feature selection and importance learning. Gray \etal \cite{gray2008viewpoint} used AdaBoost to select features. Schwartz \cite{schwartz2009learning} assigned weights to features with Partial Least Squares (PLS). Liu \etal \cite{liu2012person} developed an unsupervised approach to learn bottom-up feature importance, and adaptively weight features. 
Instead of globally weighting features across all the pedestrian images, our approach adaptively weights features based on individual person pairs to be matched, since different persons have different salient regions.

Visual features suffer from a range of variations across camera views. Feature transforms are learned to improve the invariance to cross-view transforms.
Prosser \etal \cite{prosser2008multi} learned the Cumulative Brightness Transfer Function to handle color transforms. Avraham \etal \cite{avraham2012learning, avraham2014learning} learned both implicit and explicit transforms of visual features. 
Li and Wang \cite{liWcvpr13} learned a mixture of cross-view transforms and projected features into a common space for alignment. Rather than learning feature transforms for specific camera view settings, our approach flexibly handle the cross-view variations by performing a constrained patch matching technique, which can be generalize to any disjoint camera-view transition.

Some works explored higher level features \cite{vaquero2009attribute, layne2012person,layne2012towards, liu2012attribute, zhao2014learning, li2014person} to assist person re-identification. Vaquero \etal \cite{vaquero2009attribute} first introduced mid-level facial attributes in human recognition. Layne \etal \cite{layne2012person,layne2012towards} proposed 15 human attributes for person re-identification. Song \etal \cite{liu2012attribute} used human attributes to match persons with Bayesian decision. 
saliency distribution can also be considered as one kind of high-level features.

\begin{figure*}[!ht]
\centering
\includegraphics[width=0.95\linewidth]{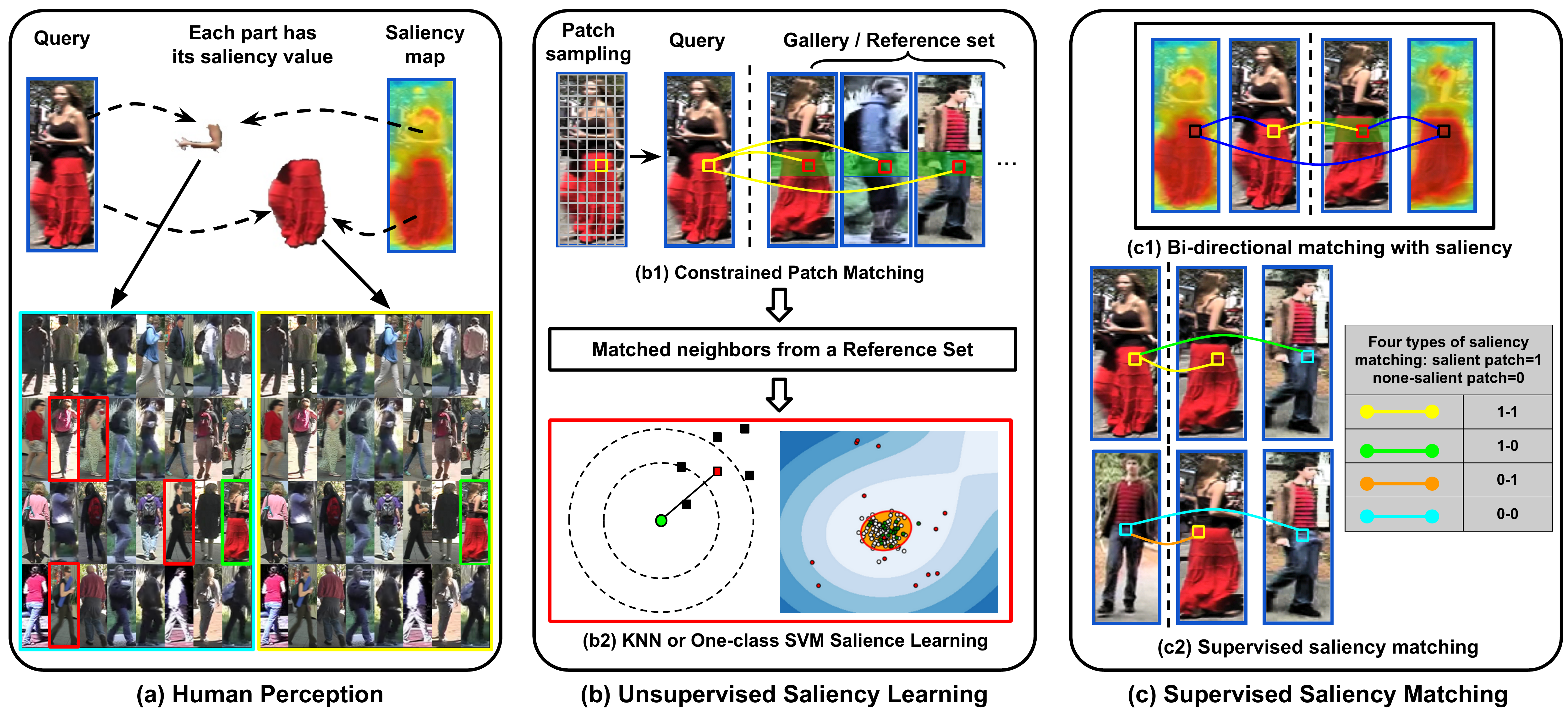}
\caption{\small{Diagram of our novel framework of human saliency learning and matching for person re-identification.} }
\label{fig:overview}
\end{figure*}

\subsection{Rank and Metric Learning}

Given a query image, an image of the same person is expected to have a high rank on the candidate list after matching. 
Prosser \emph{et al}. \cite{prosser2010person} formulated person re-identification as a ranking problem, and learned global feature weights with RankSVM. Wu \etal \cite{wu2011optimizing} introduced rank-loss optimization to improve accuracy in re-identification. Loy \etal \cite{loy2013person} exploited unlabeled gallery data to propagate labels to query instances with a manifold ranking model. Liu \etal \cite{liu2013pop} presented a man-in-loop method to allow users to quickly refine ranking result. 
In this paper, we employ structural RankSVM \cite{joachims2009cutting}, which considers ranking difference. 

Many research works \cite{zheng2011person, dikmen2011pedestrian, hirzer2012dense, hirzer2012relaxed, li2012human, liu2012person, mignon2012pcca, kostinger2012large, li2013learning,pedagadi2013local} focused on optimizing distance metrics for matching persons. Zheng \etal \cite{zheng2011person} learned the metric by maximizing the likelihood of true matches to have a smaller distance than that of a wrongly matched pair. 
Dikmen \etal \cite{dikmen2011pedestrian} proposed to learn a Mahalanobis distance that is optimal for k-nearest neighbor classification by using a maximum margin formulation. Mignon and Jurie \cite{mignon2012pcca} learned a joint projection for dimension reduction, satisfying distance constraints added by image pairs. 
Li \etal \cite{li2013learning} proposed to learn a decision function for matching, which jointly models a distance metric and a locally adaptive thresholding rule. Pedagadi \etal \cite{pedagadi2013local} employed Local Fisher Discriminant Analysis to learn a distance metric. 
Above learned metrics are based on subtraction of misaligned feature vectors, which causes significant information loss and errors. Our approach handles feature misalignment through patch matching.

\subsection{Human saliency vs. General Image saliency}

General image saliency has been well studied \cite{itti1998model, hou2012image, goferman2012context, judd2009learning, li2011co}. In the context of person re-identification, human saliency is different than general image saliency in the way of drawing visual attentions. With the aim to improve the performance of re-identification, human saliency is a considered as visual patterns of distinguishing a person from others, while general saliency draws visual attention within a single image to capture salient foreground objects from background.


\section{Method Overview}
The diagram of the proposed saliency learning and matching framework is shown in Figure \ref{fig:overview}. Section \ref{sec:reidsaliency} conducts a user study to estimate human saliency based on human perception in the person re-identification task. We investigate the discriminative power of different body regions in identifying a target person from a gallery set. The saliency of each local region of a query image is quantitatively estimated by measuring the averaged number of trails that human labelers find the target person only based on that region of the query image. An illustration is shown in Figure \ref{fig:overview} (a).  The red and green bounding boxes indicate incorrect and correct targets chosen by the labeler from the gallery. The red skirt has higher saliency and causes fewer failure trails compared with the arm. Our result shows that subjects can recognize a query person only based on a small salient part without looking at the whole image. Salient regions vary on different persons.

An unsupervised approach for saliency learning is proposed in Section \ref{ssec:saliencylearning} and illustrated in Figure \ref{fig:overview} (b). With constrained patch matching, each patch finds its matched neighbors from a reference set of training images. K-Nearest Neighbor and One-Class SVM models are employed to learn human saliency. Our experimental results show both qualitative and quantitative evaluation of the correlation between the learned saliency and human perception. With obtained human saliency, matching image pairs can be performed in unsupervised and supervised ways as described in Section \ref{ssec:saliencylearning}. For the unsupervised manner, saliency is used to bi-directionally weight patch matching similarity and penalize inconsistence of saliency distribution across camera views, as shown by the blue lines in Figure \ref{fig:overview} (c). For the supervised manner, person matching is formulated as a saliency matching problem, which considers four types of saliency matching cases, as shown in the table in Figure \ref{fig:overview} (c). Matching cost is a linear function of patch matching similarities, which is  learned with Structural RankSVM. The learned saliency matching function is used to measure similarities between images. 

\section{saliency from   Human Perception} \label{sec:reidsaliency}


We define human saliency in the context of person re-identification and estimate it by user study. The design of user study considers the following aspects.

\begin{figure} 
\centering
\includegraphics[width = 0.95\linewidth]{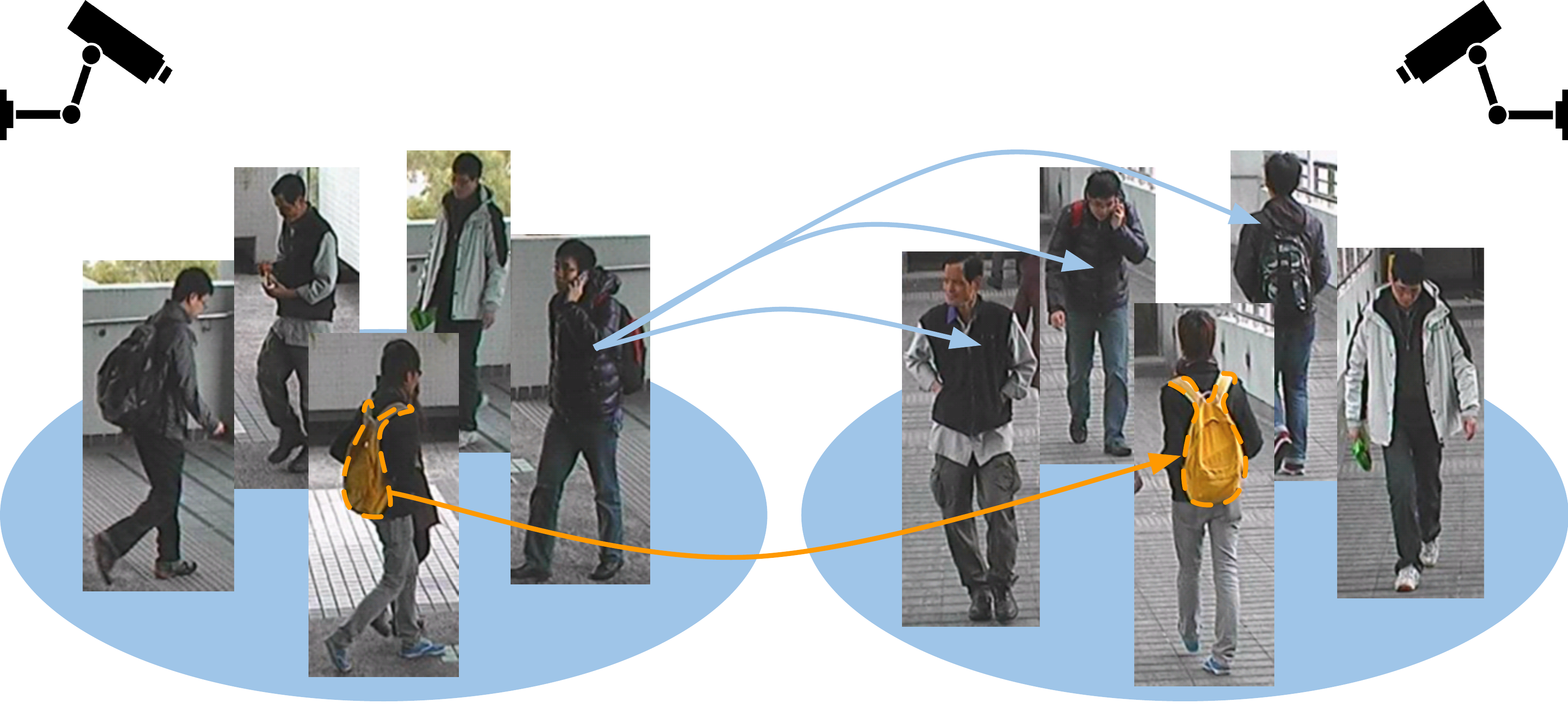} 
\caption{\small{Salient and non-salient body parts in person re-identification.}}
\label{fig:saldef}
\end{figure}


\begin{itemize}
   \item Salient body parts are those possessing unique appearance compared with a reference set.
  \item Human body, including carrying accessories, can be decomposed into parts with different saliency values.  
  \item If a body part helps subjects to quickly identify a person from other candidates across camera views, it is considered as salient. 
  \item The salient values of different parts are estimated independently to simplify analysis. Higher order saliency from combinations of body parts  could be studied in the future work.
\end{itemize}
As an example in Figure \ref{fig:saldef}, the yellow bag is a carrying accessory, which can be easily identified across camera views, while the black coat appears on many persons, and is hard to be used as a cue to recognize identity. Thus, the yellow bag has a higher saliency value. 

\begin{figure}
\centering
\includegraphics[width = 0.98\linewidth]{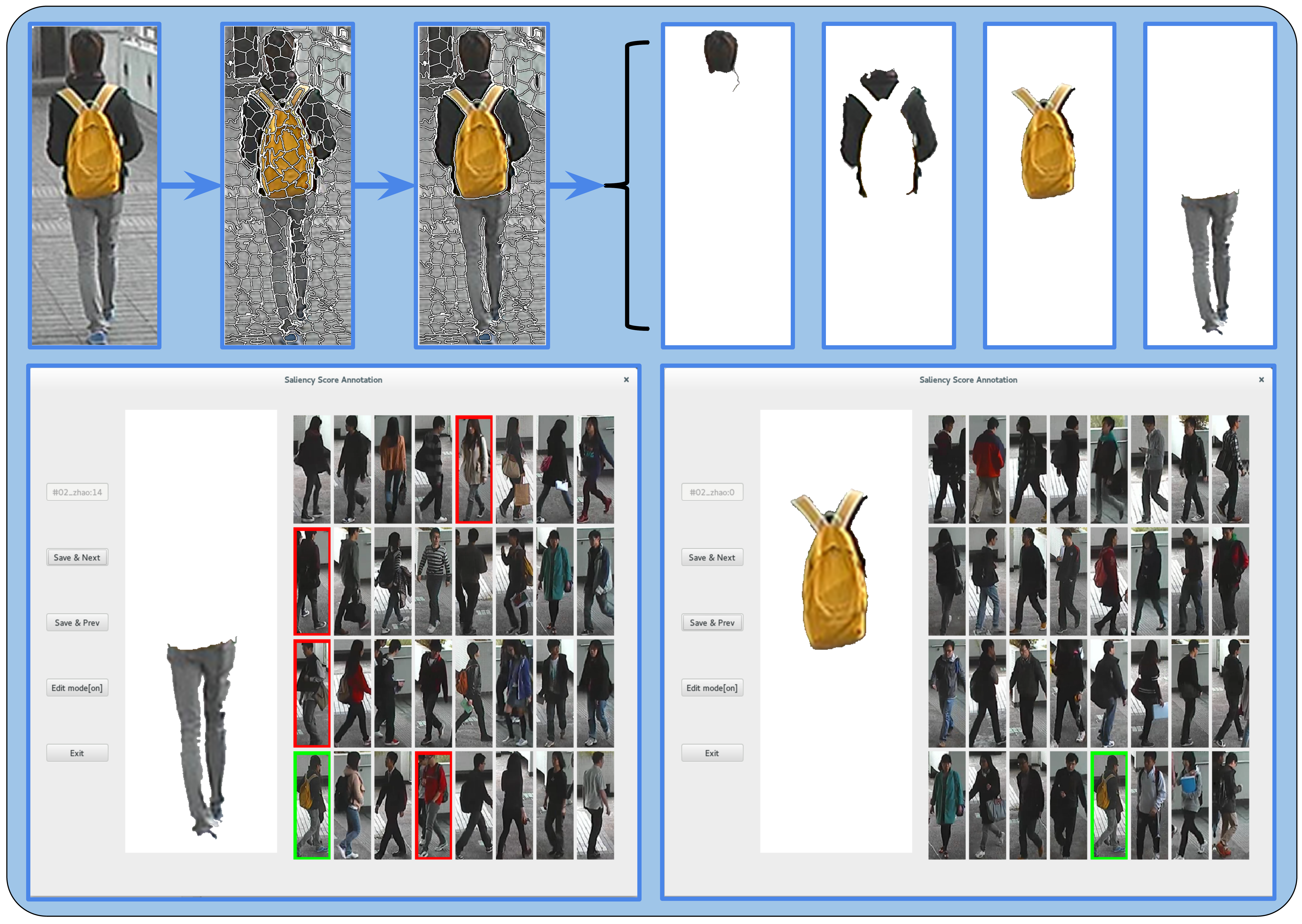} 
\caption{\small{\textbf{Flow chart of human saliency annotation.} The first row illustrates the procedure that an image is segmented into semantic body parts. The second row shows the interface of annotating human saliency.}}
\label{fig:annotation_scheme}
\end{figure}

\subsection{Human Annotation Scheme} \label{sec:gtsaliency}

Given an image, we apply superpixel segmentation \cite{achanta2012slic}, and then manually merge superpixels that are coherent in  appearance. As an example shown in the first row of Figure \ref{fig:annotation_scheme}, superpixels on the yellow bag are merged into a part. Superpixels with different semantic meanings are not merged. For example, even if the hair and jacket share similar appearance, they are annotated as two parts. Only foreground superpixels are annotated. 

A segmented body part is randomly selected and presented to a labeler for annotation. Each part is annotated for multiple times by different labelers. Their annotations are combined into a saliency value. In Figure \ref{fig:annotation_scheme}, a body part from a query image is revealed (on the left) at its original spatial location in the image while other parts are masked, and a list of 32 images randomly sampled from the gallery set are also shown (on the right) to the labeler. The true target (observed in a different camera view from the query image) is among the sampled images, but the order is randomly shuffled. The labeler is asked to select the most likely image from the list based on visual perception. The labeler is allowed to select for multiple times until the correct match is found. In the second row of Figure \ref{fig:annotation_scheme}, the red bounding boxes indicate wrong selection and the green one indicates the correct match found in the end.
A part is considered as salient if labelers try fewer times to found the target. 

Denote the $i$-th revealed part by $p_i$, and the number of trails of the $j$-th user on this part by $n_{p_i}^j$. Then the saliency value of the revealed part is estimated as
\begin{align}
    \label{eq:salscore_annotate}
    \mathbi{score}(p_i) = \exp(-\frac{m_{p_i}^2}{\sigma_{avg}^2})\exp(-\frac{s_{p_i}^2}{\sigma_{std}^2}).
\end{align}
$m_{p_i}$ and $s_{p_i}$ are the average and standard deviation of $n_{p_i}^j$ over all the labelers. $\sigma_{avg}$ and $\sigma_{std}$ are bandwidth parameters.

\begin{figure}
\centering
\includegraphics[width = 0.9\linewidth]{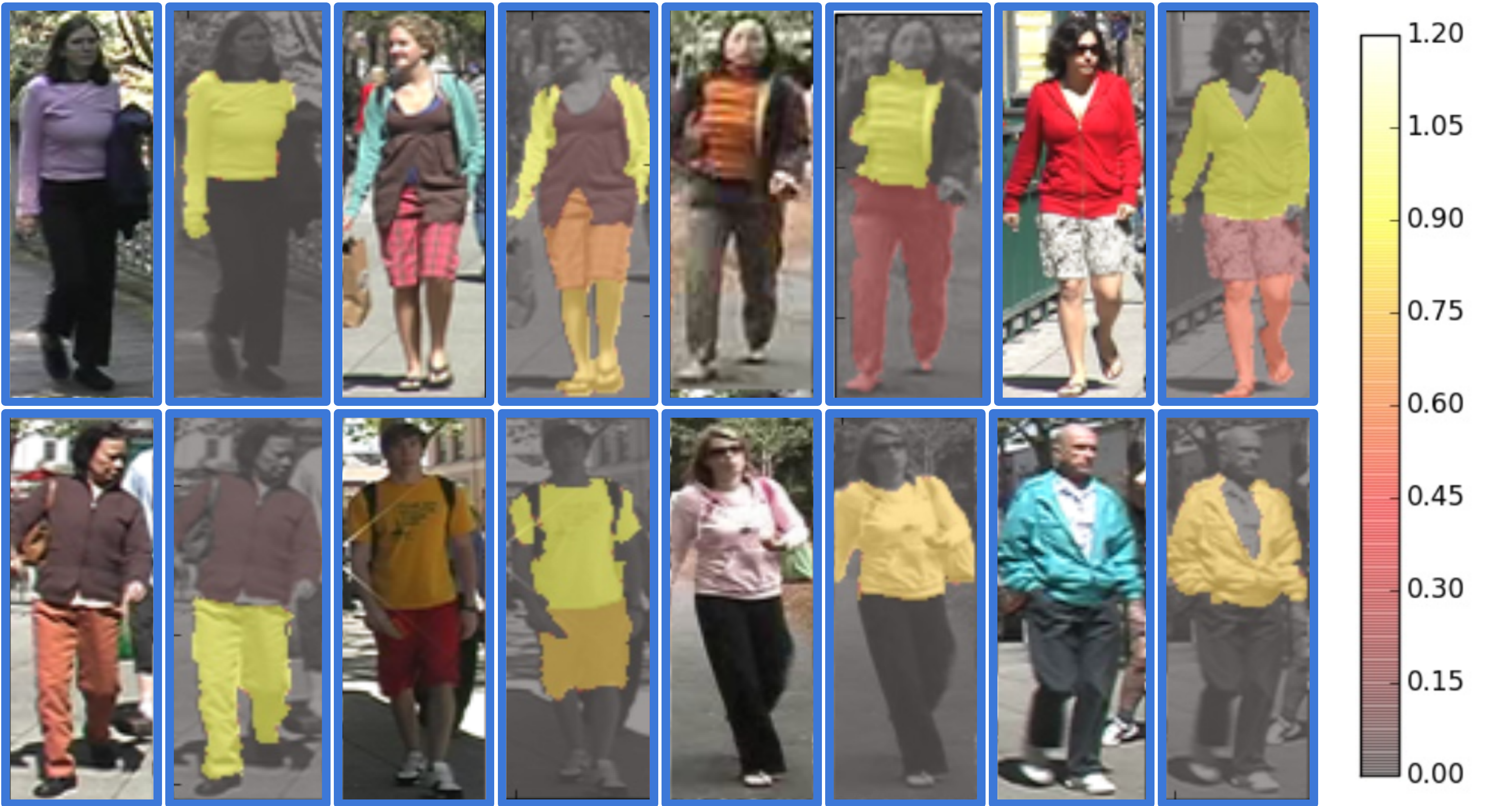} 
\caption{\small{\textbf{Examples of saliency annotation.} Each body part is annotated with a saliency value. saliency map is overlaid on the gray-level image (right). The original color image is  on the left.} }
\label{fig:annotation_result}
\end{figure}

\begin{figure}
\centering
\begin{minipage}{0.475\linewidth}
\centering 
\includegraphics[width=\linewidth]{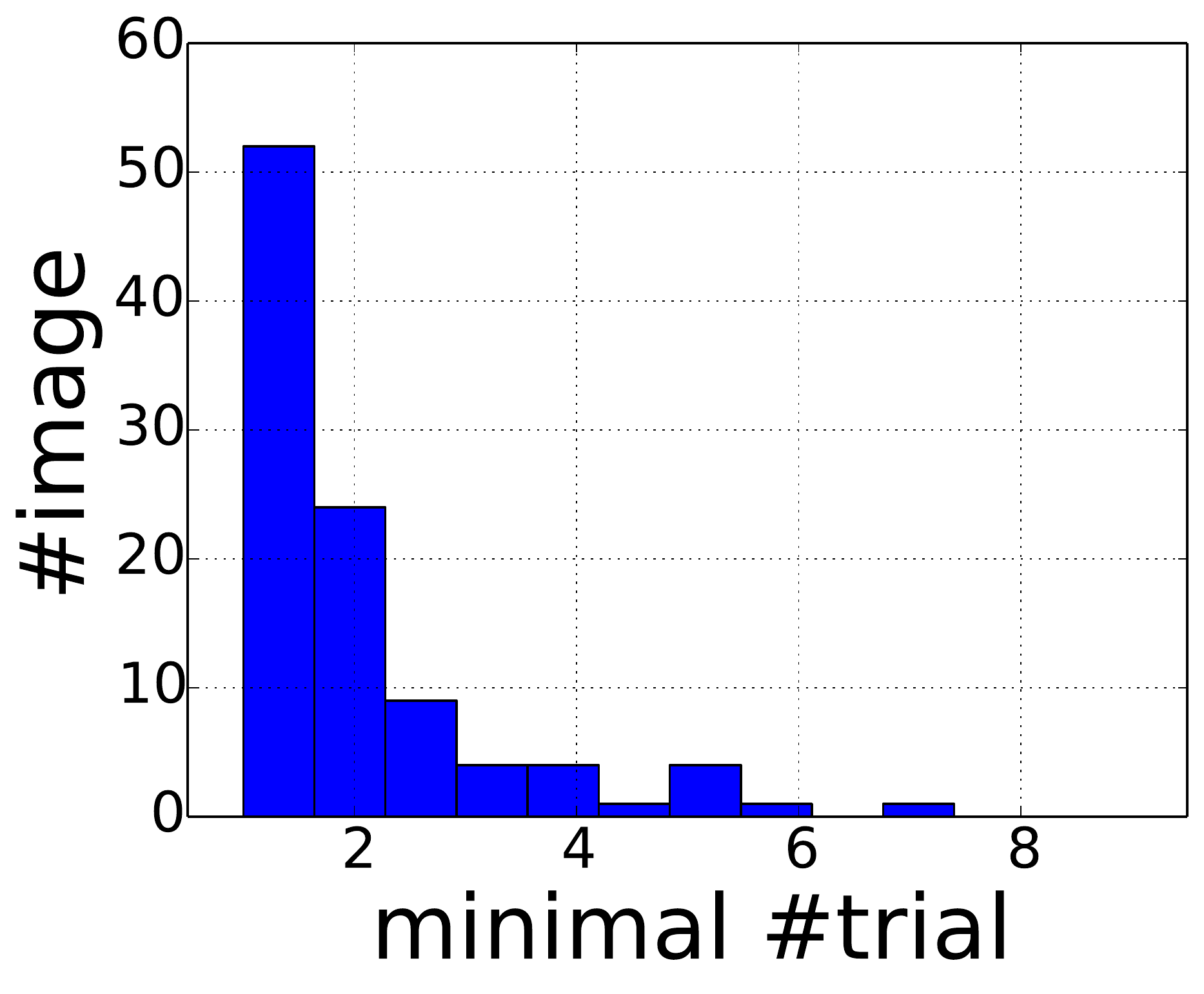} \\ (a) 
\end{minipage}
\begin{minipage}{0.48\linewidth}
\centering 
\includegraphics[width=\linewidth]{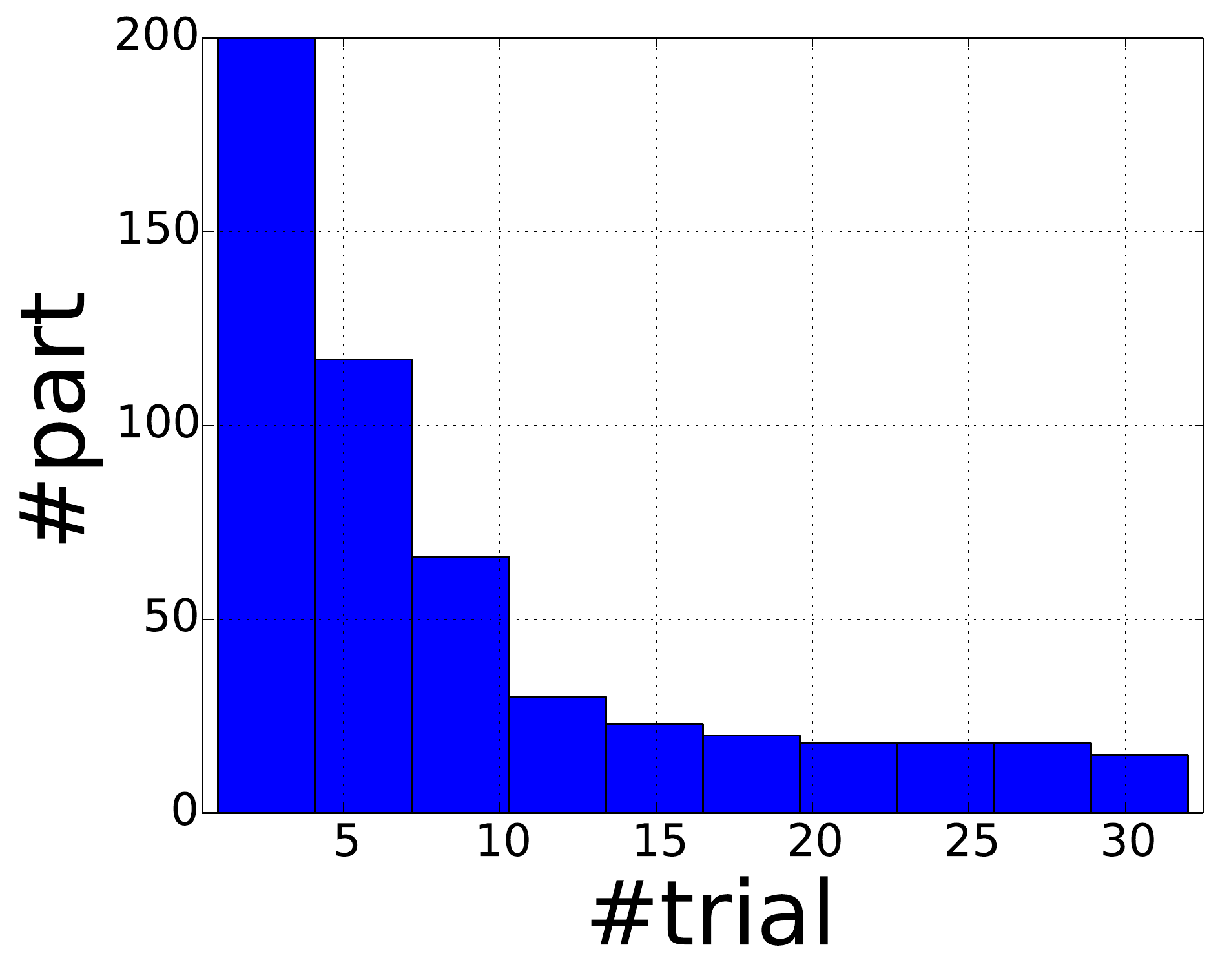} \\ (b) 
\end{minipage}
\caption{ \small{Statistics on saliency annotation. (a) Histogram on the numbers of trails used to find the targets only based on the most salient parts on query images. (b) Histogram on the numbers of trails for all the parts.} }
\label{fig:hitstats}
\end{figure}

Annotation is conducted on $524$ body parts of $100$ images from camera view $A$ of the VIPeR dataset \cite{gray2007evaluating}. 
Some examples of the annotated saliency maps are shown in Figure \ref{fig:annotation_result}. 
In order to investigate whether salient regions exist in pedestrian images, Figure \ref{fig:hitstats} (a) shows the histogram on the numbers of trails used to find the targets only based on the most salient parts on query images. It shows that more than half of the pedestrians can be recognized, if the labelers only observe the most salient part of a query image.  As comparison, Figure \ref{fig:hitstats} (b) plots the histogram on the numbers of trails for all the parts. It shows that most other body parts are not salient enough. 
The correlation between the annotated saliency and that obtained with the proposed computation model will be validated in experiments in Section \ref{ssec:results}.



\section{Human saliency Learning} \label{ssec:saliencylearning}
We propose to automatically learn human saliency in an unsupervised manner. Dense correspondence between images is first built with patch matching, and two alternative approaches (K-nearest neighbor and One-Class SVM) are proposed to estimate human saliency without using identity labels or human annotated saliency. 

\subsection{Feature Extraction}\label{sec:densefeature}
Each image is densely divided into a grid of overlapping local patches, and each patch is represented by a feature vector concatenating color histograms and SIFT features computed around its local region.

\noindent\textbf{Dense Color Histogram.}  A color histogram in LAB color space is extracted from each patch. LAB color histograms are computed on multiple downsampled scales and L2 normalized. 

\noindent\textbf{Dense SIFT.} 
We divide each patch into $4\times4$ cells, quantize the orientations of local gradients into $8$ bins, and obtain a $4\times4\times8=128$ dimentional SIFT feature vector, which is also L2 normalized. 

In our experiment, scales of pedestrian images range from 128 $\times$ 48  to 160 $\times$ 60. Patches of size $10\times10$ pixels are sampled on a dense grid with a step size $4$. $32$-bin color histograms are computed in each LAB channels, and in each channel, $3$ levels of downsampling are used with scaling factors $0.5$, $0.75$ and $1$. SIFT features are also extracted in $3$ color channels and thus produces a $128\times3$ feature vector for each patch. In a summary, each patch is finally represented with a discriminative descriptor vector of length $32\times3\times3 + 128\times3 = 672$. We denote the combined Color-SIFT feature vector as $DenseFeats$. 

\begin{figure}
\centering
\includegraphics[width = 0.55\linewidth]{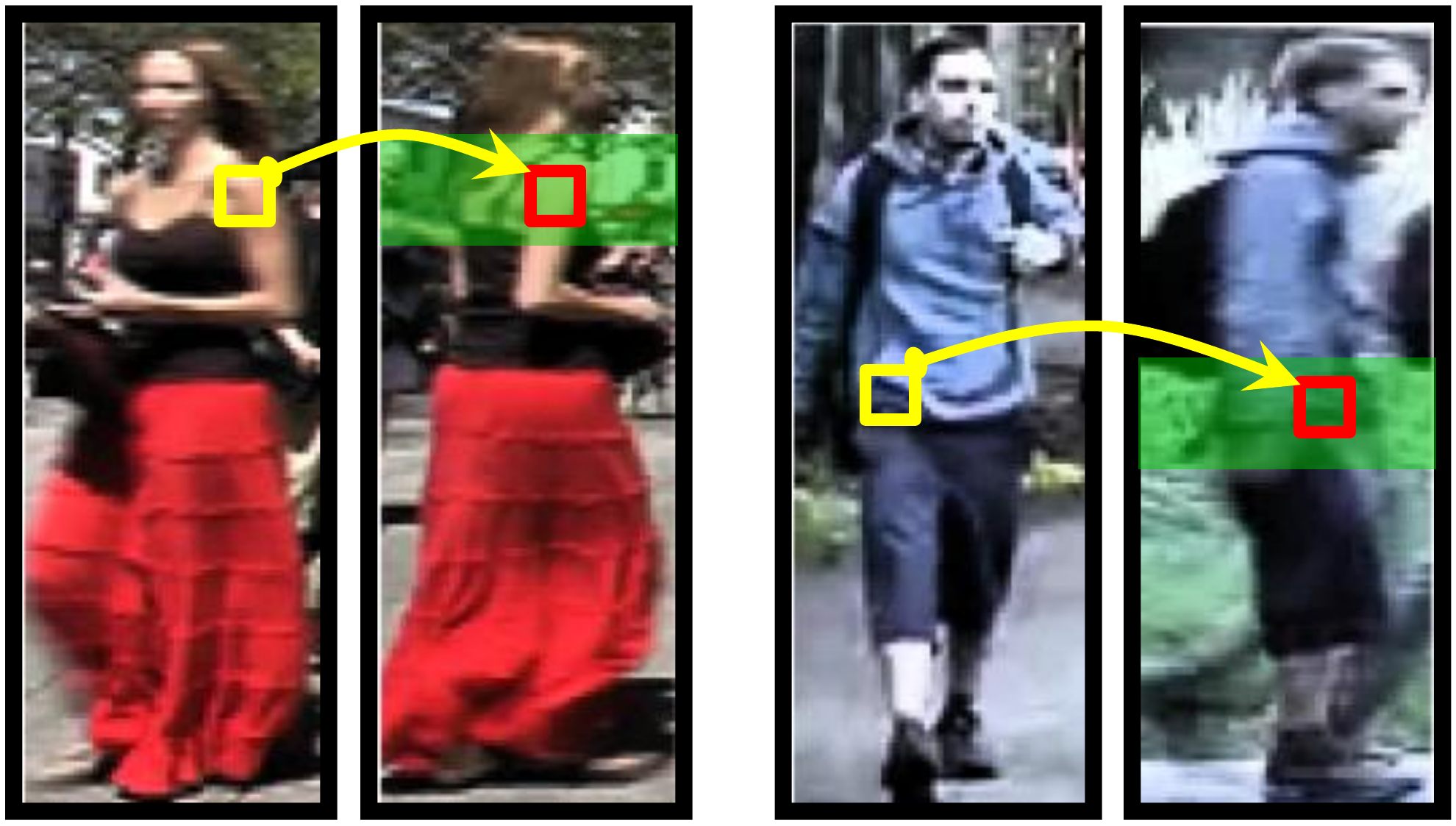}\\ 
\vspace{0.01in}
\caption{\small{Illustration of adjacency constrained search. Green region represents the adjacency constrained search set of patch in yellow box. Patch in red box is the target match.}}
\vspace{-0.15in}
    \label{fig:adjacency}
\end{figure}

\subsection{Dense Correspondence}
To deal with misalignment, we build dense correspondence between images by adjacency constrained search. 
DenseFeats features of a pedestrian image is represented as $X^{A, u} = \{\mathbf{x}^{A, u}_{m, n} ~|~ m=1\hdots,M, ~n=1\hdots,N\}$, where $(A, u)$ denotes the $u$-th image in camera $A$, $(m, n)$ denotes the patch centered at the $m$-th row and the $n$-th column of this image, and $\mathbf{x}^{A, u}_{m, n}$ is the dense Color-SIFT feature vector of the patch. A natural baseline is to compute image similarity with concatenated patch features, 
\begin{align}
  \label{eq:cp_densefeats}
  \mathbi{sim}_{DenseFeats}(X^{A, u}, X^{B, v}) = \ssum_{\substack{i=1, \hdots,M\\j=1,\hdots,N}} s(\mathbf{x}_{i, j}^{A, u}, \mathbf{x}_{i, j}^{B, v}),
\end{align}
where 
\begin{align}
\label{eq:patchsimscore}
s(\mathbf{x}_{i, j}^{A, u}, \mathbf{x}_{i, j}^{B, v}) = exp(-\frac{d(\mathbf{x}_{i, j}^{A, u}, \mathbf{x}_{i, j}^{B, v})^2}{2\sigma^2}),
\end{align}  
is the similarity between two patch features, $d(\cdot)$ is the Euclidean distance, and $\sigma$ is a bandwidth parameter. 

\textbf{Adjacency Searching.} $\mathbi{sim}_{DenseFeats}$ does not consider misalignment. We propose adjacency constrained searching to allow flexible matching among patches in image pairs. When the patches are matched with those from another image, patches in the same row have the same search set, denoted as 
\begin{align}
\mathcal{S}(\mathbf{x}^{A, u}_{m, n}, X^{B, v}) = \{\mathbf{x}^{B, v}_{i, j} ~|~ i=m, ~j = 1, \hdots, N\}.
\end{align}
$\mathcal{S}(\mathbf{x}^{A, u}_{m, n}, X^{B, v})$ restricts the search set in $X^{B, v}$ within the $m$-th row. However, bounding boxes produced by a human detector are not always well aligned, and also uncontrolled human pose variations exist. We relax the horizontal constraint to have a larger search range:
\begin{align}
\label{eq:range}
\mathcal{\hat{S}}(\mathbf{x}^{A, u}_{m, n}, X^{B, v}) = \{\mathbf{x}^{B, v}_{i, j} ~|~ i\in\mathcal{N}(m), ~j = 1, \hdots, N\},
\end{align}
where $\mathcal{N}(m) = \{\max(0, m-l), \hdots, m, \hdots, \min(m+l, M)\}$. $l$ defines the size of the relaxed adjacent vertical space. 
Less relaxed search space cannot well tolerate the spatial variation while more relaxed search space increases the chance of matching different body parts. $l=2$ is chosen in our setting.

\begin{figure}
\centering
\begin{minipage}{0.15\linewidth}
\centering
\includegraphics[width = \textwidth]{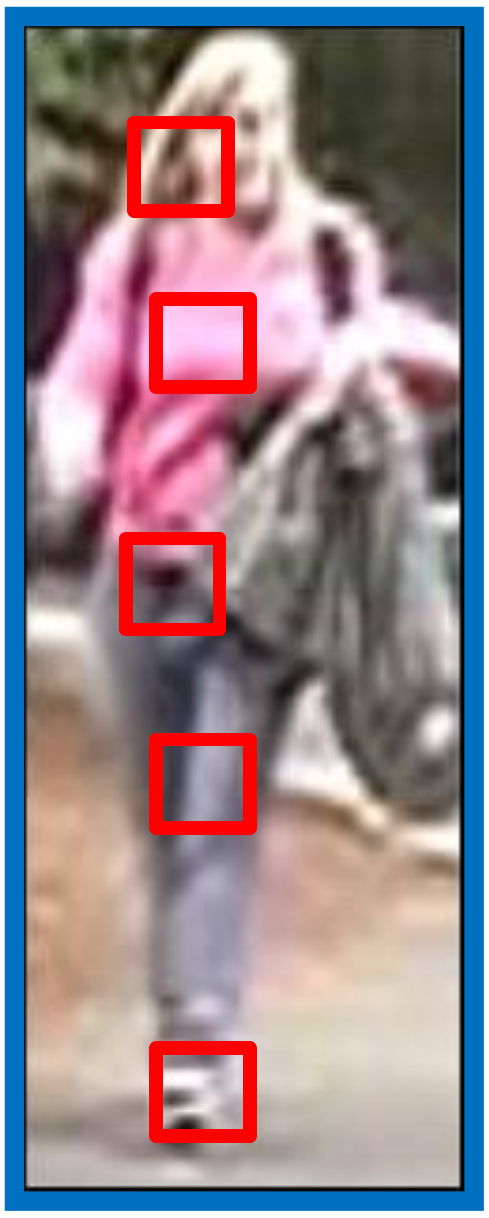} \\ \vspace{0.08in}(a)
\end{minipage}
\begin{minipage}{0.5\linewidth}
\centering
\includegraphics[width = \textwidth]{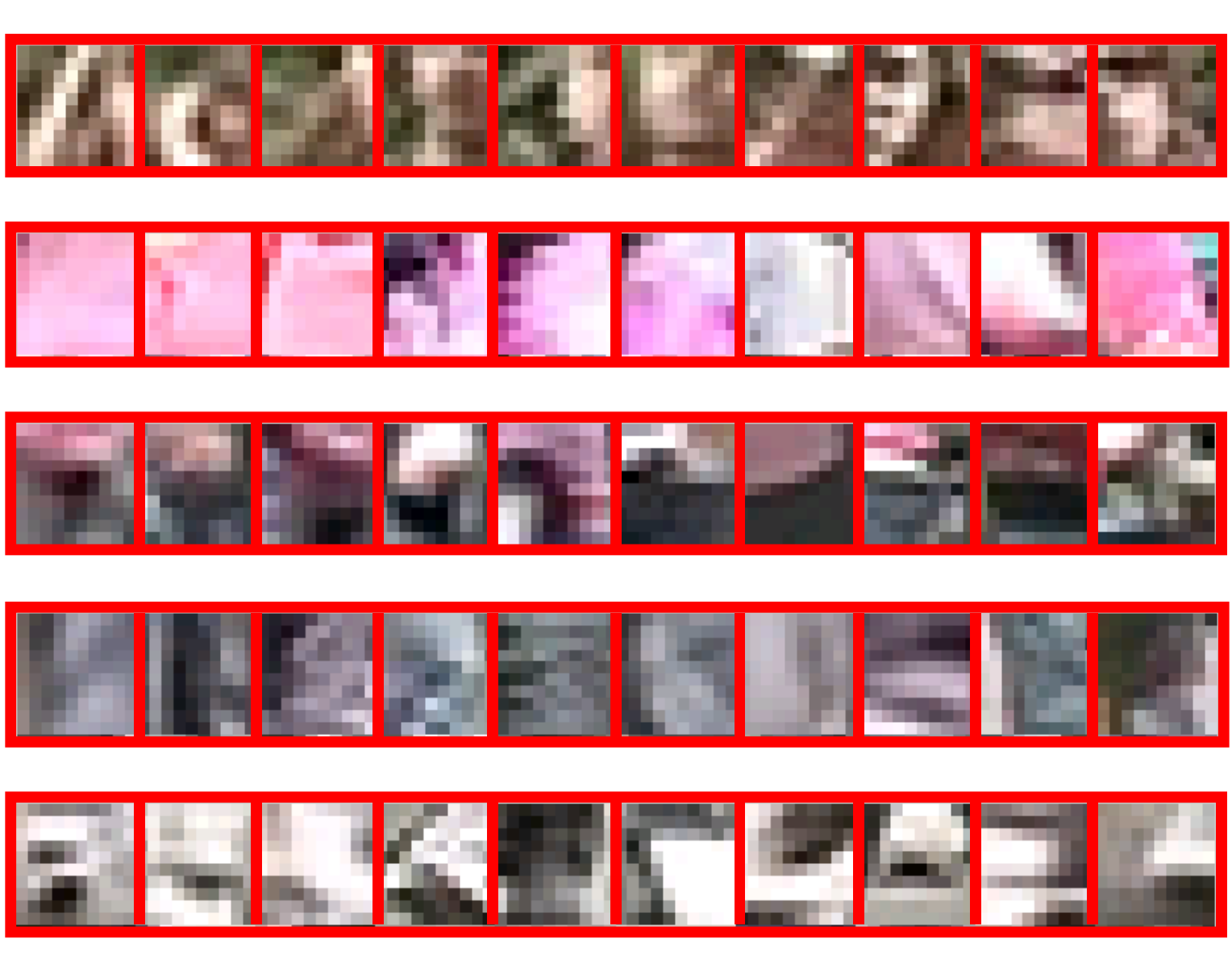} \\ (b)
\end{minipage}
\vspace{0.05in}
\caption{\small{\textbf{Examples of adjacency search.} (a) A test image from the VIPeR dataset. Local patches are densely sampled, and five exemplar patches on different body parts are shown in red boxes. (b) One nearest neighbor from each reference image is returned by adjacency search for each patch on the left, and then $N$ nearest neighbors from $N$ reference images are sorted. The top ten nearest neighbor patches are shown. Note that the ten nearest neighbors are from ten different images.}}
\label{fig:dcsearchviper_cvpr}
\end{figure}

We perform the nearest neighbor search for each $\mathbf{x}^{A, p}_{m, n}$ in its search set $\mathcal{\hat{S}}(\mathbf{x}^{A, p}_{m, n}, X^{B, q})$.
For each patch $\mathbf{x}^{A, p}_{m, n}$, a nearest neighbor is sought from its search set in every image within a reference set. The adjacency constrained search process is illustrated in Figure \ref{fig:adjacency}, 
and some visually similar patches returned by the discriminative adjacency constrained search are shown in Figure \ref{fig:dcsearchviper_cvpr}.

\subsection{Unsupervised saliency Learning}

\subsubsection{K-Nearest Neighbor (KNN) saliency} \label{sec:knnsal}
Byers \etal \cite{byers1998nearest} found the KNN distances can be used for clutter removal. Since human saliency detection shares a similar goal as abnormality detection, KNN should also be viable in finding human saliency. By searching for the K-nearest neighbors of a test patch in the set of matched patches obtained with dense correspondence, KNN is adapted to the re-identification problem. saliency score of the test patch is computed with the KNN distance. 



\begin{figure}
\centering
\centering
\begin{minipage}{0.8\linewidth}
\centering
\includegraphics[width = \textwidth]{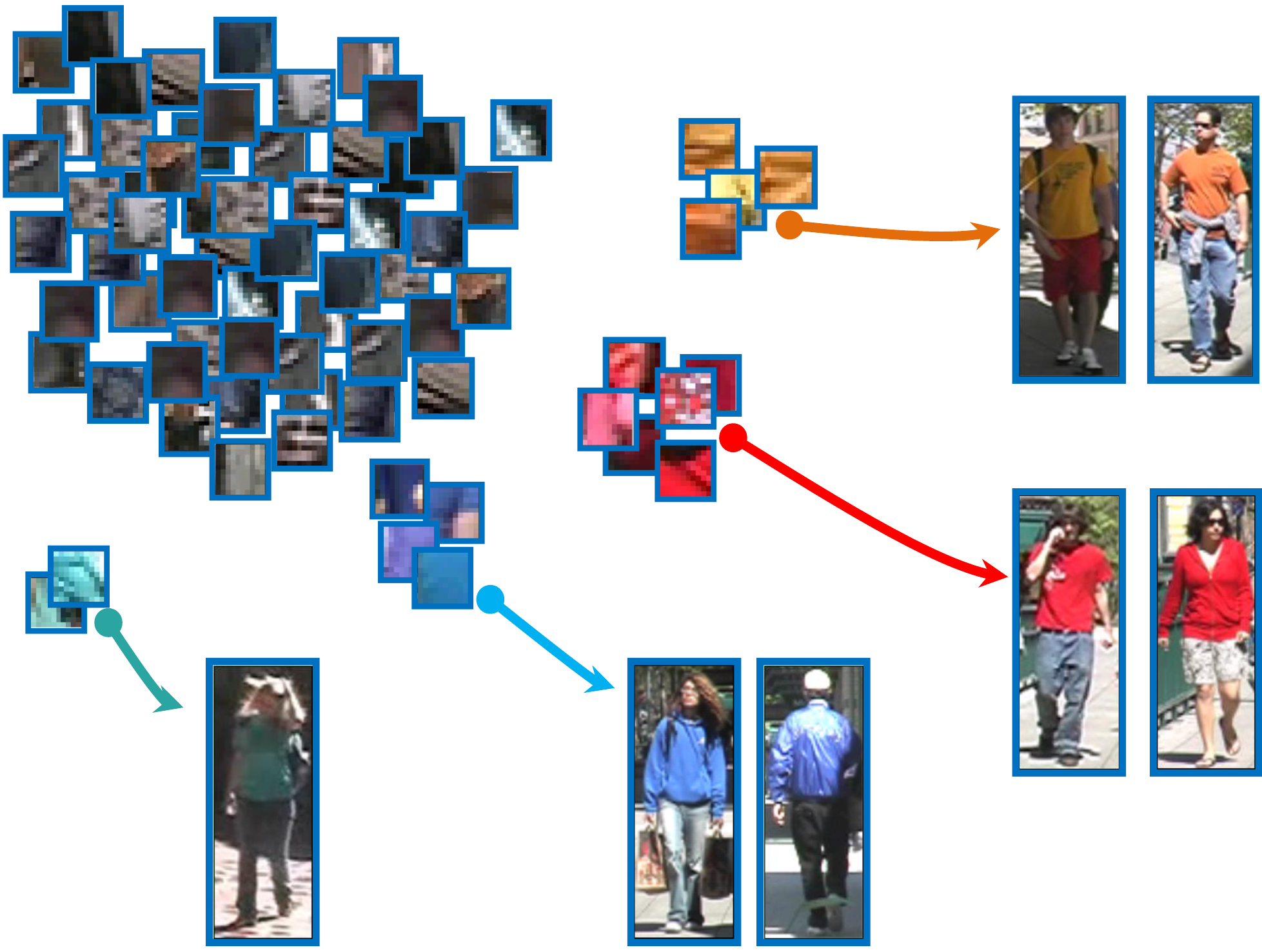}
\end{minipage}
\vspace{0.05in}
\caption{\small{\textbf{Illustration of salient patch distribution.} Salient patches are distributed far way from other pathes.} } 
\label{fig:saliency}
\vspace{-0.15in}
\end{figure}

We denote the number of images in the reference set by $N_r$. 
After building dense correspondences between a test image and reference images, a nearest neighbor (NN) set of size $N_r$ is obtained for every patch $\mathbf{x}^{A, u}_{m, n}$, 
\begin{align} \label{eq:nnset}
X_{NN}(\mathbf{x}^{A, u}_{m, n}) &= \{\mathbf{x} ~|~ \argmin_{\mathbf{x}^{B, v}_{i, j}} d(\mathbf{x}^{A, u}_{m, n}, ~\mathbf{x}^{B, v}_{i, j}), \nonumber\\
 & \mathbf{x}^{B, v}_{i, j}\in\mathcal{\hat{S}}(\mathbf{x}^{A, u}_{m, n}, X^{B, v}), ~v = 1, 2, ..., N_r\}
\end{align}
%
The KNN distances between $\mathbf{x}^{A, u}_{m, n}$ and its nearest neighbors in $X_{NN}(\mathbf{x}^{A, u}_{m, n})$ are used as the saliency score:
\begin{align} \label{eq:knnscore}
\mathbi{score}_{knn}(\mathbf{x}^{A, u}_{m, n}) = d_k(X_{NN}(\mathbf{x}^{A, u}_{m, n})),
\end{align} 
where $d_k$ denotes the distance of the $k$-th nearest neighbor. Salient patches only find a limited number ($k = \alpha_k N_r$) of visually similar neighbors, as shown in Figure \ref{fig:saliency}, and then $\mathbi{score}_{knn}(x^{A, p}_{m, n})$ is expected to be large. $0 < \alpha_k < 1$ is a proportion parameter reflecting our expectation on the statistical distribution of salient patches. 

\noindent\textbf{Choosing  $\mathbf{k}$.} The goal of saliency detection for person re-identification is to identify parts with unique appearance. 
We set $\alpha_k=0.5$ with an empirical assumption that a patch is considered to have unique appearance such that more than half of the people in the reference set do not share similar patches with it. $N_r$ reference images are randomly sampled from training set in our experiments. Enlarging the reference dataset will not deteriorate saliency detection, because saliency is defined in the statistical sense. It is robust as long as the distribution of the reference dataset well reflects the test scenario.  


\subsubsection{One-Class SVM saliency} \label{sec:ocsvmsal}
One-class SVM \cite{heller2003one} has been widely used for outlier detection. The basic idea is to use a hypersphere to describe data in the feature space and put most of the data into the hypersphere. It is formulated as an objective function: 
\begin{align}
\min_{R\in\mathbb{R}, \xi\in\mathbb{R}^l, c\in F} R^2+\frac{1}{vl}\sum_i\xi_i, \qquad\qquad\qquad\\
s.t. \|\Phi(\mathbf{x}_i)-c\|^2 \leq R^2 + \xi_i, ~~\forall i\in\{1, ...l\}: \xi_i \geq 0, \nonumber
\end{align}
where $\Phi(\mathbf{x}_i)$ is the multi-dimensional feature vector of $i$-th training sample, $l$ is the number of training samples, $R$ and $c$ are the radius and center of the hypersphere, and $v\in[0, 1]$ is a trade-off parameter. The goal is to keep the hypersphere as small as possible and include most of the training data. It can be solved in a dual form by QP optimization \cite{chen2001one}. The decision function is: 
\begin{align}
f(\mathbf{x}) = R^2 - \|\Phi(\mathbf{x})-c\|^2,
\end{align}
\begin{align}
\|\Phi(\mathbf{x})-c\|^2 = k(\mathbf{x}, \mathbf{x}) - 2\sum_i\alpha_ik(\mathbf{x}_i, \mathbf{x}) + \sum_{i, j}\alpha_i\alpha_j k(\mathbf{x}_i, \mathbf{x}_j), 
\nonumber
\end{align}
where $\alpha_i$ and $\alpha_j$ are the parameters for each constraint in the dual problem. We use the radius basis function (RBF) $K(\mathbf{x}, \mathbf{y}) = \exp\{-\|\mathbf{x}-\mathbf{y}\|^2/2\sigma^2\}$ as kernel to deal with high-dimensional, non-linear, and multi-mode distributions. As shown in \cite{chen2001one}, the decision function of kernel One-class SVM can well capture the density and modality of the feature distribution. As an alternative to the KNN saliency alogrithm (Section \ref{sec:knnsal}) without requiring the choice of $K$, saliency score is defined in terms of kernel One-class SVM decision function: 
\begin{align}\label{eq:ocsvmscore}
\mathbi{score}_{ocsvm}(\mathbf{x}^{A, u}_{m, n}) = ~d(\mathbf{x}^{A, u}_{m, n}, \mathbf{x}^*), ~~~\quad\\
\mathbf{x}^* ~= \argmax_{\mathbf{x}\in X_{NN}(\mathbf{x}^{A, u}_{m, n})}f(\mathbf{x}). \nonumber
\end{align} 
Our experiments show very similar results in person re-identification with both saliency detection methods. $\mathbi{score}_{ocsvm}$ performs slightly better than $\mathbi{score}_{knn}$ in some circumstances. 
\begin{algorithm}[t]
\caption{Human saliency learning.}
\begin{algorithmic}[1] \label{alg:saliency}
\REQUIRE image $X^{A, u}$ and a reference image set $\mathcal{R} = \{X^{B, v}, ~v = 1,\hdots,N_r\}$
\ENSURE saliency probability map $P(l^{A, u}_{m, n} = 1~|~\mathbf{x}^{A, u}_{m, n})$
  \FOR{each patch $\mathbf{x}^{A, u}_{m, n} \in X$}
  \STATE compute $X_{NN}(\mathbf{x}^{A, u}_{m, n})$ with Eq. (\ref{eq:nnset})
  \STATE compute $\mathbi{score}_{opt}(\mathbf{x}^{A, u}_{m, n}), ~opt \in \{knn, ocsvm\}$ with Eq. (\ref{eq:knnscore}) or Eq. (\ref{eq:ocsvmscore})
  \STATE compute $P(l^{A, u}_{m, n} = 1~|~\mathbf{x}^{A, u}_{m, n})$ with Eq. (\ref{eq:salientprob})
  \ENDFOR
\end{algorithmic}
\end{algorithm}
The probability of $\mathbf{x}^{A, u}_{m, n}$ being a salient patch is 
\small
\begin{align} \label{eq:salientprob}
  P(l^{A, u}_{m, n} = 1 ~|~ \mathbf{x}^{A, u}_{m, n}) = 1 - \exp(-\mathbi{score}_{opt}(\mathbf{x}^{A, u}_{m, n})^2/\sigma_0^2),
\end{align}
\normalsize
where $opt \in \{knn, ~ocsvm\}$. The human saliency learning is summarized in Algorithm \ref{alg:saliency}. 

\section{saliency Matching}
One of our main contributions is to match human images based on  saliency probability map. It is based on our observation that person in different camera views shows consistence in saliency probability maps, as shown in Figure \ref{fig:intro2}. Since matching is applied to arbitrary image pairs, we omit the image index in notation for concise clarity, ~\ie~ change $X^{A, u}$ to $X^A$, $X^{B, v}$ to $X^B$, $\mathbf{x}^{A, u}_{m, n}$ to $\mathbf{x}^A_{p_i}$ and $\mathbf{x}^{B, v}_{i, j}$ to $\mathbf{x}^B_{p'_i}$. 
 $p_i$ is the patch index in image $X^A$ and $p'_i$ is the corresponding matched patch index in image $X^B$ produced by dense correspondence. We denote the dense correspondence between $X^{A}$ and $X^{B}$ as $P=\{(p_i, p'_i)\}_{i=1, \hdots, MN}$.

\subsection{Bi-directional Weighted Matching}
We first denote the method of only using patch matching without saliency information as $PatMatch$, and the image similarity is expressed as
\begin{align}
  \label{eq:cp_patmatch}
  \mathbi{sim}_{PatMatch}(X^{A}, X^B) = \sum_{(p_i, p'_i) \in P} s(\mathbf{x}^A_{p_i}, \mathbf{x}^B_{p'_i}).
\end{align}
 $s(\mathbf{x}^A_{p_i}, \mathbf{x}^B_{p'_i})$ is the visual similarity between patches. 
Searching for the best matched image in the gallery is formulated as finding the maximal similarity score. 
\begin{align}
  v^* = \argmax_{v} \mathbi{sim}(X^{A, u}, X^{B, v})
\end{align}
A bi-directional weighted matching is designed to incorporate saliency information. We denote this method as saliency guided dense correspondence ($SDC$), as illustrated in Figure \ref{fig:overview}(c1), and the similarity between two images is computed as
\small 
\begin{align}
  \label{eq:cp_sdc}
  \mathbi{sim}_{SDC_{opt}} = \sum_{(p_i, p'_i) \in P} \frac{\mathbi{score}_{opt}(\mathbf{x}^A_{p_i})\cdot s(\mathbf{x}^A_{p_i}, \mathbf{x}^B_{p'_i}) \cdot\mathbi{score}_{opt}(\mathbf{x}^B_{p'_i})}{\alpha_{sdc} + |\mathbi{score}_{opt}(\mathbf{x}^A_{p_i}) - \mathbi{score}_{opt}(\mathbf{x}^B_{p'_i})|},
\end{align} 
\normalsize  
where $\alpha_{sdc}$ is a parameter representing a base penalty. Intuitively, large saliency scores in both matched patches are expected to enhance the similarity score of matched patches. In another aspect, images of the same person would be more likely to have similar saliency distributions than those of different persons, so the difference in saliency score can be used as a penalty to the similarity score. We set $\alpha_{sdc}=1$ in experiments. 

\subsection{Unified saliency Matching}
To incorporate saliency into matching, we introduce $L^A = \{l^A_{p_i}~|~l^A_{p_i}\in \{0, 1\}\}$ and $L^B = \{l^B_{p'_i} ~|~l^B_{p'_i}\in\{0, 1\}\}$ as saliency labels for all the patches in image $X^A$ and $X^B$ respectively. If all the saliency labels are known, we can perform person matching by computing the saliency matching score as follows:
\begin{align} \label{eq:salmatch}
&f_z(X^A, X^B, L^A, L^B; P, Z) = \\
&\qquad  \sum_{(p_i, p'_i)\in P} \Big\{z_{p_i, 1}l^A_{p_i}l^B_{p'_i} + z_{p_i, 2}l^A_{p_i}(1 - l^B_{p'_i}) \nonumber\\
&\qquad+ z_{p_i, 3} (1 - l^A_{p_i})l^B_{p'_i} + z_{p_i, 4} (1 - l^A_{p_i})(1 - l^B_{p'_i}) \Big\}\nonumber,
\end{align}  
where $Z = \{z_{p_i, k}\}_{i=1,\hdots,MN, ~k=1,2,3,4}$ are the matching scores for four different saliency matching results at one local patch. $z_{p_i, k}$ is not a constant for all the patches. Instead, it depends on the spatial location $p_i$. For example, the score of matching patches on the background should be different than those on legs.  $z_{p_i, k}$ also depends on the visual similarity between patches $\mathbf{x}^A_{p_i}$ and patch $\mathbf{x}^B_{p'_i}$. Instead of directly using the Euclidean distance $d(\mathbf{x}^A_{p_i}, \mathbf{x}^B_{p'_i})$, we convert it to similarity to reduce the side effect in summation of very large distances in incorrect matching, caused by misalignment, occlusion, or background clutters.  

Therefore, we define the matching score $z_{p_i, k}$ as a linear function of the similarity as follows, 
\begin{align} \label{eq:linear}
   z_{p_i, k} = \alpha_{p_i, k}\cdot s(\mathbf{x}^A_{p_i}, \mathbf{x}^B_{p'_i}) + \beta_{p_i, k}.
\end{align} 
 $\alpha_{p_i, k}$ and $\beta_{p_i, k}$ are weighting parameters. Thus Eq. (\ref{eq:salmatch}) considers both saliency matching and visual similarity.  

Since the saliency labels $l^A_{p_i}$ and $l^B_{p'_i}$ in Eq. (\ref{eq:salmatch}) are hidden variables, 
they can be marginalized by computing the expectation of the saliency matching score as 
\begin{align} \label{eq:phi1}
&f^*(X^A, X^B; P, Z) \nonumber \\
&=\sum_{L^A, L^B} f_z(X^A, X^B, L^A, L^B; P, Z)p(L^A, L^B|X^A, X^B) \nonumber\\
&=\sum_{(p_i, p'_i)\in P} \sum_{k=1}^4\Big[\alpha_{p_i, k}\cdot s(\mathbf{x}^A_{p_i}, \mathbf{x}^B_{p'_i}) + \beta_{p_i, k}\Big]c_{p_i,k}(\mathbf{x}^A_{p_i}, \mathbf{x}^B_{p'_i}),
\end{align}
where $c_{p_i,k}(\mathbf{x}^A_{p_i}, \mathbf{x}^B_{p'_i})$ is the probabilistic saliency matching cost depending on saliency probabilities $P(l^A_{p_i} = 1~|~\mathbf{x}^A_{p_i})$ and $P(l^B_{p'_i} = 1~|~\mathbf{x}^B_{p'_i})$ given in Eq. (\ref{eq:salientprob}), 
\begin{align}
&c_{p_i,k}(\mathbf{x}^A_{p_i}, \mathbf{x}^B_{p'_i}) \\
&=\left\{
\begin{tabular}{lr} 
   $p(l^A_{p_i} = 1~|~\mathbf{x}^A_{p_i})\cdot p(l^B_{p'_i} = 1~|~\mathbf{x}^B_{p'_i}),$ & $k = 1$, \\
   $p(l^A_{p_i} = 1~|~\mathbf{x}^A_{p_i})\cdot p(l^B_{p'_i} = 0~|~\mathbf{x}^B_{p'_i}),$ & $k = 2$, \\
   $p(l^A_{p_i} = 0~|~\mathbf{x}^A_{p_i})\cdot p(l^B_{p'_i} = 1~|~\mathbf{x}^B_{p'_i}),$ & $k = 3$, \\
   $p(l^A_{p_i} = 0~|~\mathbf{x}^A_{p_i})\cdot p(l^B_{p'_i} = 0~|~\mathbf{x}^B_{p'_i}),$ & $k = 4$. \\
\end{tabular} \right. \nonumber
\end{align}
To better formulate this learning problem, we extract out all the weighting parameters in Eq. (\ref{eq:phi1}) as $\mathbf{w}$, and have
\begin{align}\label{eq:salmatchscore} 
   f^*(X^A, X^B; P, Z) &= \mathbf{w}^\textrm{T}\Phi(X^A, X^B; P) \qquad\quad\\
   &= \sum_{(p_i, p'_i)\in P} \mathbf{w}^\textrm{T}_{p_i} \phi(\mathbf{x}^A_{p_i}, \mathbf{x}^B_{p'_i}) \nonumber,
\end{align} 
where
\begin{align}
\label{eq:phi_image}
\Phi(X^A, X^B; P) &= [\phi(\mathbf{x}^A_{p_1}, \mathbf{x}^B_{p'_1})^\textrm{T},\hdots,\phi(\mathbf{x}^A_{p_{MN}}, \mathbf{x}^B_{p'_{MN}})^\textrm{T}]^\textrm{T}, \nonumber\\ 
\mathbf{w} &= [\mathbf{w}_{p_1},\hdots,\mathbf{w}_{p_{MN}}]^\textrm{T} \nonumber,\\
\mathbf{w}_{p_i} &= [\{\alpha_{p_i, k}\}_{k=1,2,3,4}, ~\{\beta_{p_i, k}\}_{k=1,2,3,4}].
\end{align}
$\Phi(X^A, X^B; P)$ is the feature map describing the matching between $X^A$ and $X^B$. For each patch $p_i$, the matching feature $\phi(\mathbf{x}^A_{p_i}, \mathbf{x}^B_{p'_i})$ is an eight dimensional vector:
\begin{align}\label{eq:featmap}
&\phi(\mathbf{x}^A_{p_i}, \mathbf{x}^B_{p'_i}) = \\ 
&\quad\left[
\begin{tabular}{r} 
$s(\mathbf{x}^A_{p_i}, \mathbf{x}^B_{p'_i})\cdot p(l^A_{p_i} = 1~|~\mathbf{x}^A_{p_i})\cdot p(l^B_{p'_i} = 1~|~\mathbf{x}^B_{p'_i})$ \\
 $s(\mathbf{x}^A_{p_i}, \mathbf{x}^B_{p'_i})\cdot p(l^A_{p_i} = 1~|~\mathbf{x}^A_{p_i})\cdot p(l^B_{p'_i} = 0~|~\mathbf{x}^B_{p'_i})$ \\
 $s(\mathbf{x}^A_{p_i}, \mathbf{x}^B_{p'_i})\cdot p(l^A_{p_i} = 0~|~\mathbf{x}^A_{p_i})\cdot p(l^B_{p'_i} = 1~|~\mathbf{x}^B_{p'_i})$  \\
 $s(\mathbf{x}^A_{p_i}, \mathbf{x}^B_{p'_i})\cdot p(l^A_{p_i} = 0~|~\mathbf{x}^A_{p_i})\cdot p(l^B_{p'_i} = 0~|~\mathbf{x}^B_{p'_i})$ \\
 $p(l^A_{p_i} = 1~|~\mathbf{x}^A_{p_i})\cdot p(l^B_{p'_i} = 1~|~\mathbf{x}^B_{p'_i})$ \\
 $p(l^A_{p_i} = 1~|~\mathbf{x}^A_{p_i})\cdot p(l^B_{p'_i} = 0~|~\mathbf{x}^B_{p'_i})$ \\
 $p(l^A_{p_i} = 0~|~\mathbf{x}^A_{p_i})\cdot p(l^B_{p'_i} = 1~|~\mathbf{x}^B_{p'_i})$ \\
 $p(l^A_{p_i} = 0~|~\mathbf{x}^A_{p_i})\cdot p(l^B_{p'_i} = 0~|~\mathbf{x}^B_{p'_i})$
\end{tabular} \right] \nonumber.
\end{align} 

As shown in Eq. (\ref{eq:featmap}), the pairwise feature map $\Phi(X^A, X^B; P)$ combines the saliency probability map with appearance matching similarities. For each query image $X^A$, the images in the gallery are ranked according to the expectations of saliency matching scores in Eq. (\ref{eq:phi1}). There are three advantages of matching with human saliency : (1) the human saliency probability distribution is more invariant than other features in different camera views; (2) because the saliency probability map is built based on dense correspondence, it inherits the property of tolerating spatial variation; and (3) it can be weighted by visual similarity to improve the performance of person re-identification. We will present the details in next section by formulating the person re-identification problem with $\Phi(X^A, X^B; P)$ in the structural RankSVM framework. 

\subsection{Ranking by Partial Order} \label{sec:rankpo}
We cast person re-identification as a ranking problem for supervised training. The ranking problem will be solved by finding an optimal partial order,  mathematically defined in Eq. (\ref{eq:partial_order})(\ref{eq:partial_order_feat})(\ref{eq:opt_partial_order}). Given a dataset of pedestrian images,  $\mathcal{D}^A = \{X^{A, u}, id^{A, u}\}^U_{u=1}$ from camera view $A$ and $\mathcal{D}^B = \{X^{B, v}, id^{B, v}\}^V_{v=1}$ from camera view $B$, where $X^{A, u}$ is the $u$-th image, $id^{A, u}$ is its identity label, and $U$ is the total number of images in $\mathcal{D}^A$. Similar notations apply for variables of camera view $B$. Each image $X^{A, u}$ has its relevant images (same identity) and irrelevant images (different identities) in dataset $\mathcal{D}^B$. 
Our goal is to learn the weight parameters $\mathbf{w}$ that order relevant gallery images before irrelevant ones. For the image $X^{A, u}$, we rank the relevant images before irrelevant ones, but no information of the orders within relevant images or irrelevant ones is provided. 
The partial order $\mathbf{y}^{A, u}$ is denoted as,
\begin{align}
\label{eq:partial_order}
\mathbf{y}^{A, u} = \{y^{A, u}_{v, v'}\},
\quad y^{A, u}_{v, v'} = \left\{ 
\begin{tabular}{lr} 
   $+1$ & $X^{B, v} \prec X^{B, v'}$ ,\\
   $-1$ & $X^{B, v} \succ X^{B, v'}$ ,
\end{tabular} \right. 
\end{align}
where $X^{B, v} \prec X^{B, v'}$ ($X^{B, v} \succ X^{B, v'}$) represents that $X^{B, v}$ is ranked before (after) $X^{B, v'}$ in partial order $\mathbf{y}^{A, u}$.

The partial order feature \cite{joachims2005support, mcfee2010metric} is appropriate for our goal and can encode the difference between relevant pairs and irrelevant pairs with only partial orders. 
The partial order feature for image $X^{A, u}$ is formulated as, 
\small
\begin{align}
\label{eq:partial_order_feat}
\Psi_{po}(X^{A, u}, \mathbf{y}^{A, u}; \{X^{B, v}\}^V_{v=1}, \{P^{u, v}\}^V_{v=1}) = \quad\qquad\qquad\qquad\quad\nonumber\\
\ssum_{\substack{X^{B, v}\in S_{X^{A, u}}^+ \\ X^{B, v'}\in S_{X^{A, u}}^-}} y^{A, u}_{v, v'}\frac{\Phi(X^{A, u}, X^{B, v}; P^{u, v}) - \Phi(X^{A, u}, X^{B, v'}; P^{u, v'})}{|S_{X^{A, u}}^+|\cdot|S_{X^{A, u}}^-|},
\end{align} 
\normalsize
\vspace{-0.25in}
\begin{align}
S_{X^{A, u}}^+ = \{X^{B, v} ~|~ id^{B, v} = id^{A, u} \}, \qquad\qquad\\
S_{X^{A, u}}^- = \{X^{B, v} ~|~ id^{B, v} \neq id^{A, u} \}, \qquad\qquad
\end{align} 
where $\{P^{u, v}\}_{v=1}^V$ are the dense correspondences between image $X^{A, u}$ and every gallery image $X^{B, v}$, $S_{X^{A, u}}^+$ is relevant image set of $X^{A, u}$, $S_{X^{A, u}}^-$ is irrelevant image set, $\Phi(X^{A, u}, X^{B, v}; P^{u, v})$ is the feature map defined in Eq. (\ref{eq:phi_image}), and the difference vector of two feature maps $\Phi(X^{A, u}, X^{B, v}; P^{u, v}) - \Phi(X^{A, u}, X^{B, v'}; P^{u, v'})$ is added if $X^{B, v} \prec X^{B, v'}$ or subtracted otherwise. 

A partial order may correspond to multiple rankings. Our task is to find a good ranking satisfying the optimal partial order $\mathbf{y}^{A, u}_*$ that maximizes the following score function, 
\small
\begin{align}
   \label{eq:opt_partial_order}
   \mathbf{y}^{A, u}_* =
   \argmax_{\mathbf{y}^{A, u}\in\mathcal{Y}^{A, u}} ~ \mathbf{w}^\textrm{T}\Psi_{po}(X^{A, u}, \mathbf{y}^{A, u}; \{X^{B, v}\}^V_{v=1}, \{P^{u, v}\}^V_{v=1}),
\end{align}
\normalsize
where $\mathcal{Y}^{A, u}$ is the space consisting of all the possible partial orders. As discussed in \cite{joachims2005support, yue2007support},  good ranking can be obtained by sorting gallery images by $\{\mathbf{w}^\textrm{T}\Phi(X^{A, u}, X^{B, v}; P^{u, v})\}_v$ in a descending order. The remaining problem is how to learn $\mathbf{w}$. With an optimized $\mathbf{w}_*$, we denote the unified saliency matching similarity as  
\begin{align}
\label{eq:cp_salmatch}
   \mathbi{sim}_{SalMatch_{opt}}(X^A, X^B) = \mathbf{w}_*^\textrm{T}\mathbf{\Phi}(X^{A}, X^{B};P),
\end{align}
where $opt\in\{knn,~ocsvm\}$.

\subsection{Structural RankSVM Training}
We employ structural SVM to learn the weighting parameters $\mathbf{w}$. Different than many previous SVM-based approaches \cite{carterette2006learning, prosser2010person} doing optimization  over pairwise differences, structural SVM optimizes over ranking differences and can incorporate non-linear multivariate loss functions  into global optimization in SVM training. 


\vspace{0.1in}
\textbf{Objective function}. Our goal is to learn a linear model and the training is based on n-slack structural SVM \cite{joachims2009cutting}. The objective function is as follows,
\begin{align}\label{eq:obj}
   \min_{\mathbf{w}, \mathbf{\xi}}~~\frac{1}{2}\|\mathbf{w}\|^2 + C\sum_{u=1}^U \xi_u ,\qquad\qquad\qquad\quad\\
   \mathit{s.t.}~~\mathbf{w}^\textrm{T}\delta\Psi_{po}(X^{A, u}, \mathbf{y}^{A, u}, \hat{\mathbf{y}}^{A, u}; \{X^{B, v}\}^V_{v=1}, \{P^{u, v}\}^V_{v=1}) \nonumber\\
   \geq \Delta(\mathbf{y}^{A, u}, \hat{\mathbf{y}}^{A, u}) - \xi_u, \qquad\nonumber\\
    ~~\forall \hat{\mathbf{y}}^{A, u}\in \mathcal{Y}^{A, u}\diagdown\mathbf{y}^{A, u}, ~\xi_u \geq 0, ~for~ u = 1,\hdots,U, ~~\quad\nonumber
\end{align}
where $\delta\Psi_{po}$ is defined as
\begin{align}
\delta\Psi_{po}(X^{A, u}, \mathbf{y}^{A, u}, \hat{\mathbf{y}}^{A, u}; \{X^{B, v}\}^V_{v=1}, \{P^{u, v}\}^V_{v=1}) \quad\nonumber\\
 =\Psi_{po}(X^{A, u}, \mathbf{y}^{A, u}; \{X^{B, v}\}^V_{v=1}, \{P^{u, v}\}^V_{v=1}) \quad \nonumber\\
 -\Psi_{po}(X^{A, u}, \hat{\mathbf{y}}^{A, u}; \{X^{B, v}\}^V_{v=1}, \{P^{u, v}\}^V_{v=1}), 
\end{align}
$\mathbf{w}$ is the weight vector, $C$ is a parameter to balance between  margin and  training error, $\mathbf{y}^{A, u}$ is a correct partial order that ranks all correct matches before incorrect matches, and $\hat{\mathbf{y}}^{A, u}$ is an incorrect partial order that violates some of the pairwise relations, \eg a correct match is ranked after an incorrect match in $\hat{\mathbf{y}}^{A, u}$. 
The constraints in Eq. (\ref{eq:obj}) force the discriminant score of correct partial order $\mathbf{y}^{A, u}$ to be larger than that of incorrect one $\hat{\mathbf{y}}^{A, u}$ by a margin, which is determined by a loss function $\Delta(\mathbf{y}^{A, u}, \hat{\mathbf{y}}^{A, u})$ 
 and a slack variable $\xi_u$.

\textbf{AUC loss function}.  
Many loss functions can be applied in structural SVM. 
In person re-identification, we choose the ROC Area loss, which is also known as Area Under Curve (AUC) loss. It is computed from the number of swapped pairs,
\begin{align}
   N_{swap} = \{(v, v'): X^{B, v} \succ X^{B, v'} ~and~\qquad\qquad\\
   \mathbf{w}^\textrm{T}\Phi(X^{A, u}, X^{B, v};P^{u, v}) < \mathbf{w}^\textrm{T}\Phi(X^{A, u}, X^{B, v'};P^{u, v'})\} \nonumber,
\end{align} 
\ie the number of pairs of samples that are not ranked in a correct order. In the case of partial order ranking, the loss function is 
\begin{align} \label{eq:aucloss0}
   \Delta(\mathbf{y}^{A, u}, \hat{\mathbf{y}}^{A, u}) &= |N_{swap}| / |S_{X^{A, u}}^+|\cdot|S_{X^{A, u}}^-|,\\
   &= \sum_{v, v'}(1 - \hat{y}^{A, u}_{v, v'})/(2\cdot|S_{X^{A, u}}^+|\cdot|S_{X^{A, u}}^-|). \nonumber
\end{align}
We note that there are an exponential number of constraints in Eq. (\ref{eq:obj}) due to the huge dimensionality of $\mathcal{Y}^{A, u}$. Joachims \textit{et al.}  \cite{joachims2009cutting} showed that the problem could be efficiently solved by a cutting plane algorithm. In our problem, the discriminative model is learned by the structural RankSVM algorithm, and the weight vector $\mathbf{w}$ in our model means how important it is for each term in Eq. (\ref{eq:featmap}). In Eq. (\ref{eq:featmap}), $\{\alpha_{p_i, k}\}_{k=1,2,3,4}$ correspond to the first four terms based on saliency matching with visual similarity, and $\{\beta_{p_i,k}\}_{k=3,4}$ correspond to the last four terms only depending on saliency matching.

\begin{figure}
\centering 
\includegraphics[width = 0.87\linewidth]{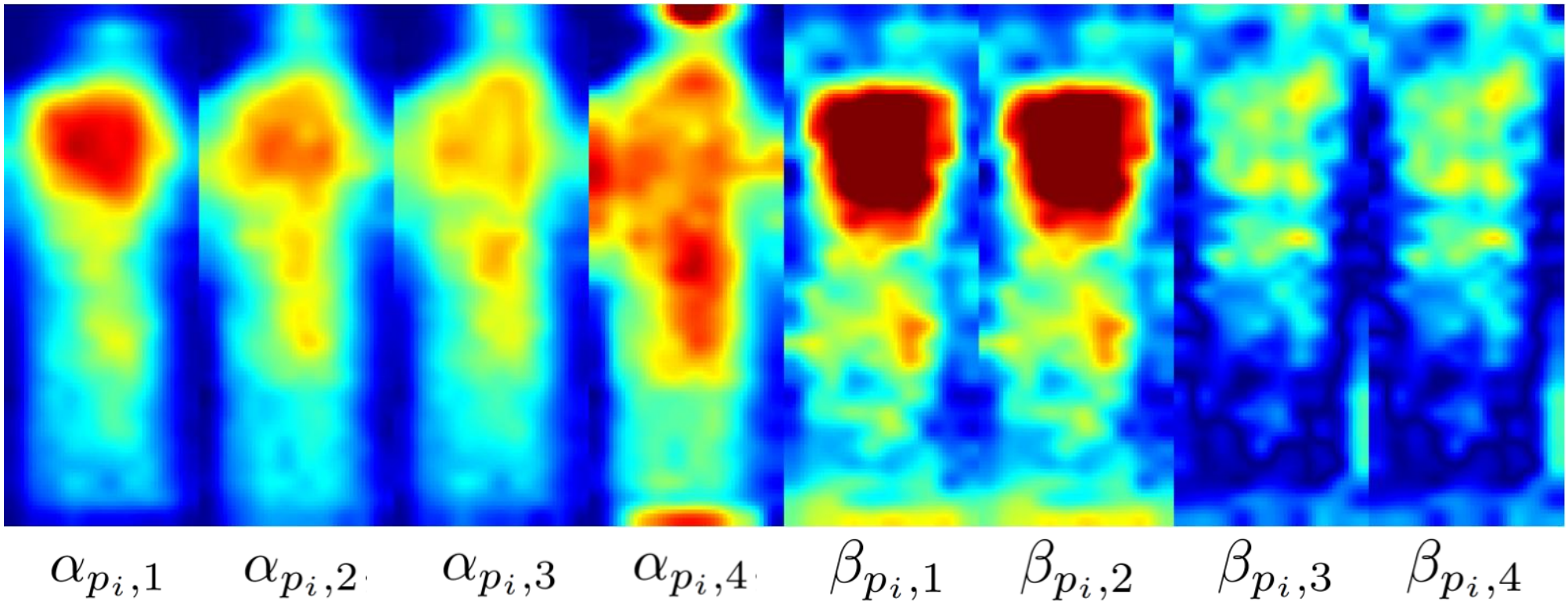}\\
    \caption{\small {We Normalize the learnt weight vector $\mathbf{w}$ to a 2-dimensional importance map for different spatial locations. Eight importance maps correspond to $\{\alpha_{p_i, k}\}_{k=1, 2, 3, 4}$ and $\{\beta_{p_i, k}\}_{k = 1, 2, 3, 4}$ in Eq. (\ref{eq:phi1}).}}
    \label{fig:modelw}
\end{figure}

We visualize the learning result of $\mathbf{w}$ in Figure \ref{fig:modelw}, and find that the first four terms in Eq. (\ref{eq:featmap}) are heavily weighted in the central part of human body which implies the importance of saliency matching based on visual similarity. $\{\beta_{p_i,k}\}_{k=1,2}$ are not relevant to visual similarity and they correspond to the two cases when $l_{p_i}^A=1$, \ie the patches on the query images are salient. It is observed that their weighting maps are highlighted on the upper body, which matches to our observation that salient patches usually appear on the upper body. $\{\beta_{p_i,k}\}_{k=3,4}$ are not relevant to visual similarity either, but they correspond to the cases when $l_{p_i}^A=0$, \ie the patches on the query images are not salient. We find that their weights are very low on the whole maps. It means that non-salient patches on query images have little effect on person re-identification if the contribution of visual similarity is not considered. 


\subsection{Combination with existing approaches}
\label{ssec:combination}
Our approach is complementary to existing approaches. In order to combine existing approaches with the matching score in Eq. (\ref{eq:salmatchscore}), 
the distance between two images can be computed as follows:
\begin{align}
\label{eq:cp_esalmatch}
   \mathbi{sim}_{eSalMatch_{opt}}(X^A, X^B) = \sum_{i}{ \mu_{i}\cdot \mathbi{sim}_{i}(X^A, X^B)}  \nonumber\\ 
   -\mu_{Sal}\cdot \mathbi{sim}_{SalMatch_{opt}}(X^A, X^B) 
\end{align}
where $\mu_{i}(>0)$ is the weight for the $i$th similarity measure, $\mu_{Sal} (> 0)$ the weight for unified saliency matching similarity. $\mathbi{sim}_i$ corresponds to the similarity measures using wHSV and MSCR in \cite{farenzena2010person} or LADF \cite{li2013learning}. In the experiment, $\{\mu_{i}\}$ are chosen the same as in \cite{farenzena2010person,li2013learning}. $\mu_{Sal}$ is fixed as $1$. 

\section{Experimental Results} \label{ssec:results}
We evaluate our approach on two public datasets, \emph{i.e}. the VIPeR dataset \cite{gray2007evaluating}, and the CUHK01 dataset \cite{li2012human}. Examples of images in the two datasets are shown in Figure \ref{fig:salmaps}. Qualitative results of saliency learning are shown, and quantitative results are reported in standard Cumulated Matching Characteristics (CMC) curves \cite{wang2007shape}.

\subsection{Datasets}

\indent\textbf{VIPeR Dataset\cite{gray2007evaluating}}. The VIPeR dataset \footnote{\url{http://vision.soe.ucsc.edu/?q=node/178}} contains images from two cameras, which were placed at many different locations in an outdoor academic environment.  Therefore, the viewpoint changes between cameras are complex. From the time it was publicly available, it has become one of the most challenging person re-identification datasets. It contains 632 pedestrian pairs, each pair contains two images of the same individual seen from different cameras. 
Most of the image pairs show viewpoint change larger than 90 degree. All images are normalized to $128\times 48$ for experiments. 

\vspace{0.1in}
\noindent\textbf{CUHK01 Dataset \cite{li2012human}}. The CUHK01 dataset\footnote{\url{http://www.ee.cuhk.edu.hk/~xgwang/CUHK_identification.html}} was also captured from two camera views in a campus environment. Images in this dataset are of higher resolution and are more suitable to show the effectiveness of saliency matching. It has 971 persons, and each person has two images from camera $A$ and the other two from camera $B$. Camera $A$ is from a frontal view and camera $B$ is from a side view. All images are normalized to $160\times 60$ for evaluations.


The CUHK01 dataset was recently built and contains more images than VIPeR ($3884 ~vs.~ 1264$). Both are very challenging datasets for person re-identification because they contain significant variations on viewpoints, poses, and illuminations, and their images are with occlusions and background clutters. 

\subsection{Evaluation Protocol} 
Our experiments on both datasets follow the evaluation protocol in \cite{gray2008viewpoint}, \emph{i.e}. we randomly partition the dataset into two even parts, 50\% for training and 50\% for testing. Images from camera $A$ are used as probe and those from camera $B$ as gallery. Each probe image is matched with every image in gallery, and the rank of correct match is obtained. Rank-$k$ matching rate is the expectation of correct match at rank $k$, and the cumulated values of matching rate at all ranks is recorded as one-trial CMC result. 
10 trials of evaluation are conducted to achieve stable statistics, and the expectation is reported. We denote our approach by $SalMatch$ for comparison.

\subsection{Evaluation on saliency Learning}

\begin{figure}
\centering
\begin{minipage}{0.52\linewidth}
\centering 
\includegraphics[width=\linewidth]{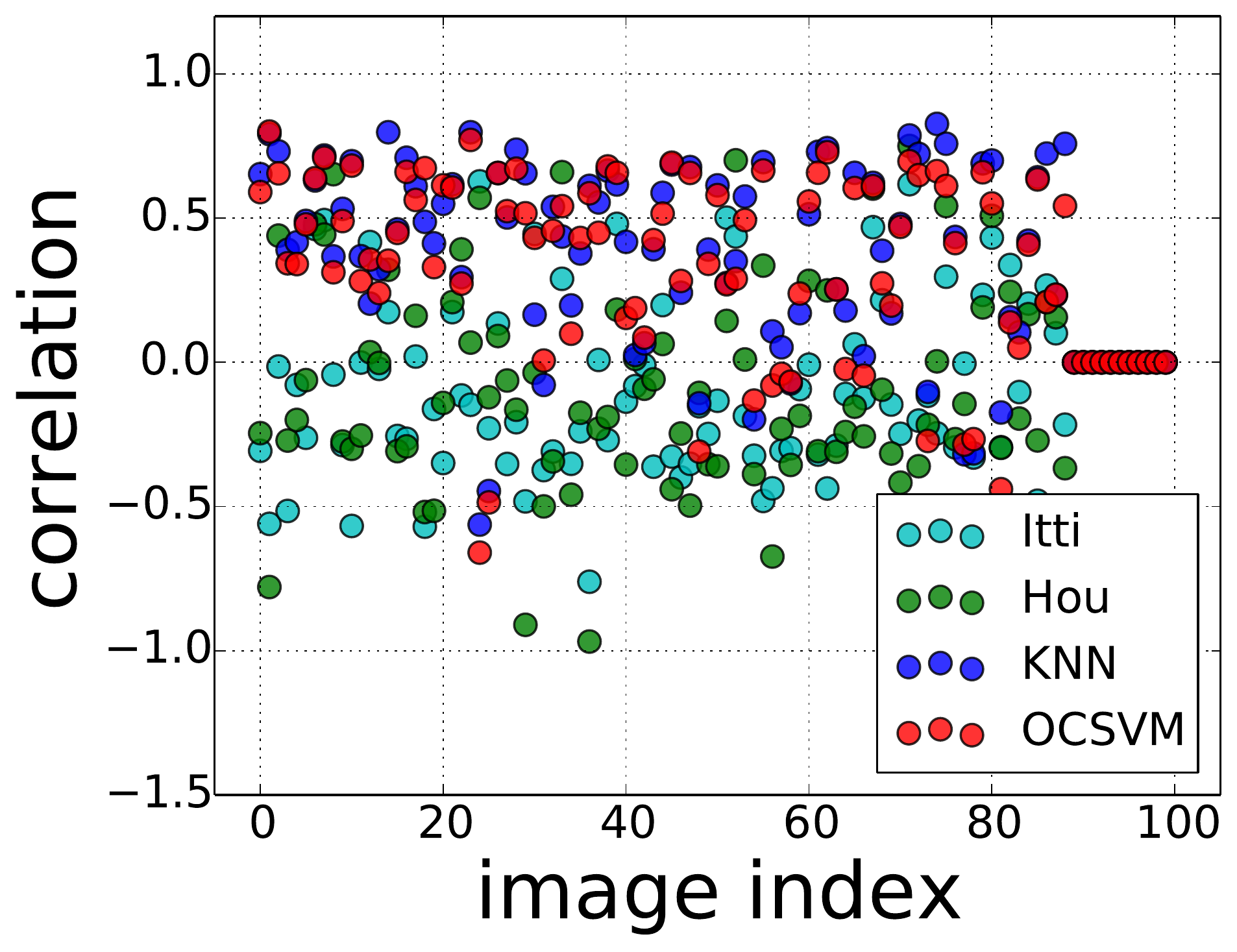} \\ (a) 
\end{minipage}
~
\begin{minipage}{0.43\linewidth}
\centering 
\vspace{0.25in}
\includegraphics[width=\linewidth]{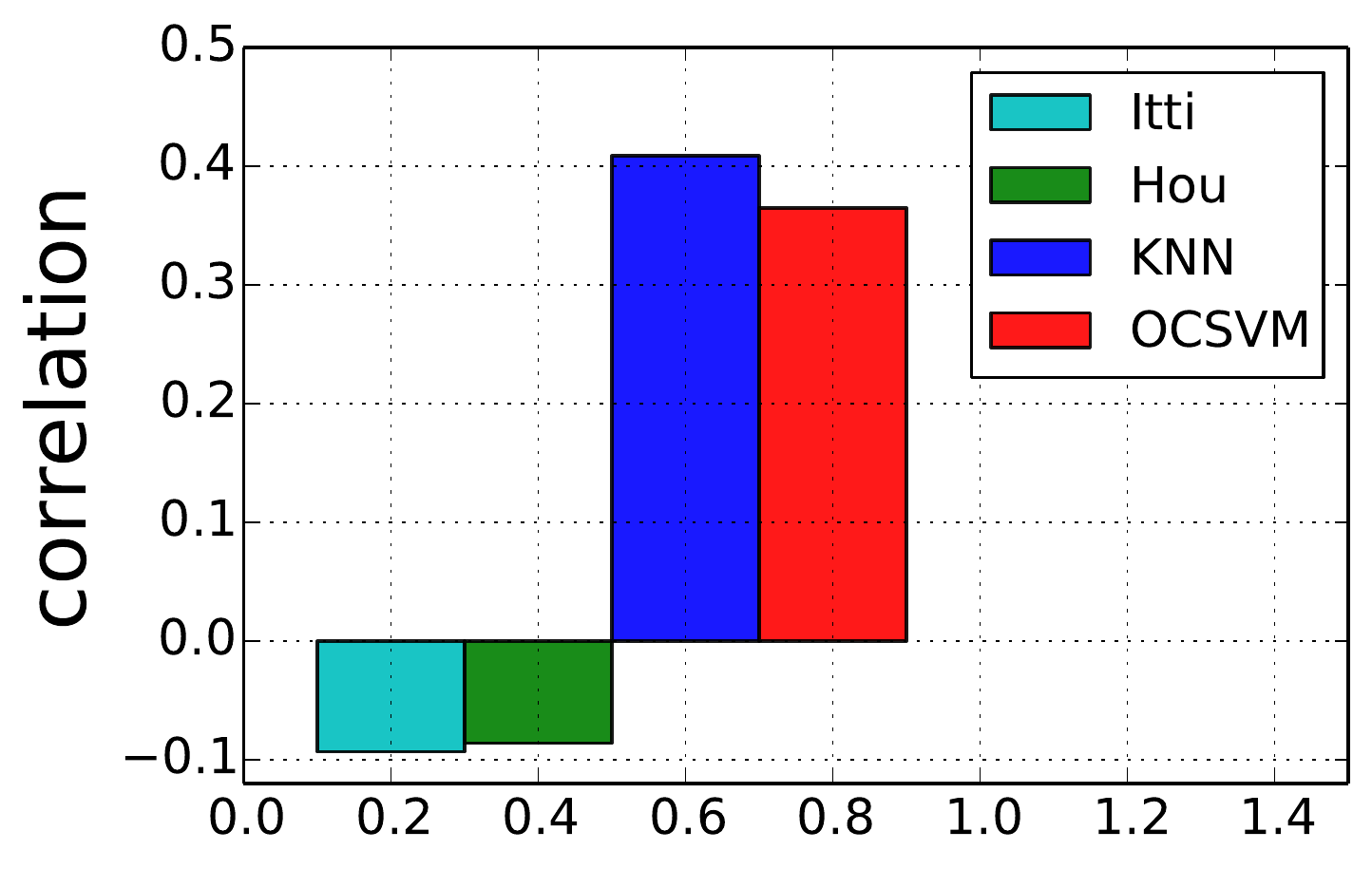} \\ \vspace{0.2in}(b) 
\end{minipage}
\caption{ \small{Correlation between automatically estimated saliency by different approaches (Itti \cite{itti1998model}, Hou \cite{hou2012image}, our KNN model and our One-Class SVM (OCSVM) model) and estimation from human perception. (a) Scatter plot of correlations over $100$ images. (b) Average correlations. }}
\label{fig:correlations}
\end{figure}

\begin{figure}
  \centering
  \includegraphics[width=0.95\linewidth]{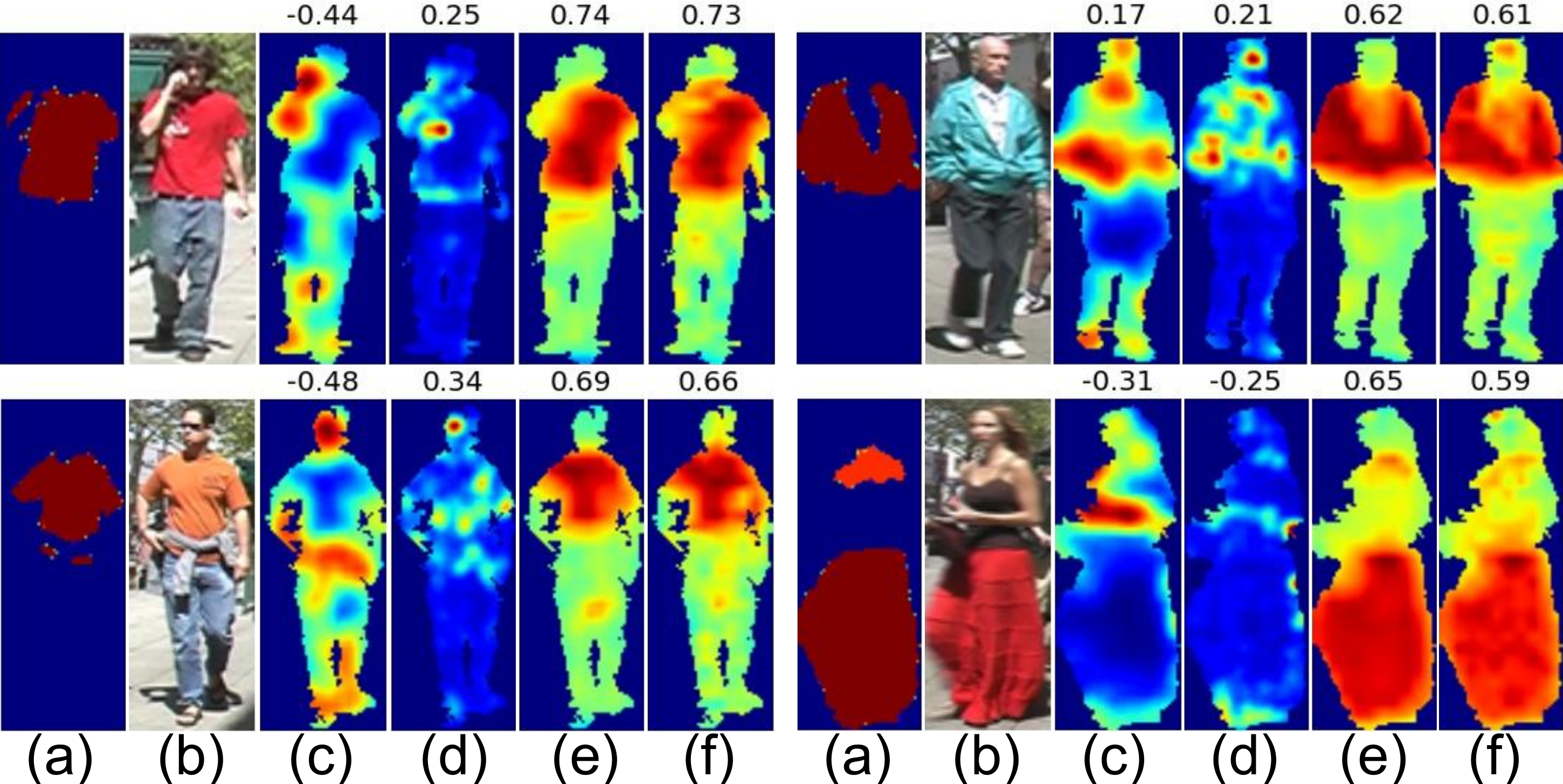} 
  \caption{\small{Examples of estimated saliency map (only body parts are shown). (a) Human saliency estimated from user study. (b) Pedestrian images. (c) and (d) are general image saliency estimated by Itti \cite{itti1998model} and Hou \cite{hou2012image}. (e) and (f) are human saliency estimated by KNN and OCSVM. Number on top of each saliency map indicates the correlation with human saliency estimated from user study.}} 
  \label{fig:correlations_example}
\end{figure}

We investigate the correlation between the human saliency estimated from human perception through user study and that automatically estimated by computation models. The computation models include those design for general image saliency (such as Itti \cite{itti1998model} and Hou \cite{hou2012image}) and our KNN and One-Class SVM (OCSVM) models specially desgined for human saliency. 
We compute the mean saliency score of each annotated body part, and the Pearson correlation between the automatically estimated saliency and estimation from human perception. Results are shown , The scatter map in Figure \ref{fig:correlations}(a) shows our learned saliency (KNN and OCSVM) has high positive correlations with human perception over the $100$ annotated images, while general image saliency (Itti and Hou) exhibits slight negative correlations. Figure \ref{fig:correlations}(b) shows averaged correlations. 
Some compared examples are shown in Figure \ref{fig:correlations_example}. The approaches for general image saliency detection can separate body parts from background. However, the identified body parts may not be effective on recognizing identities.

\begin{figure}[!t]
\centering
\includegraphics[width = 0.49\textwidth]{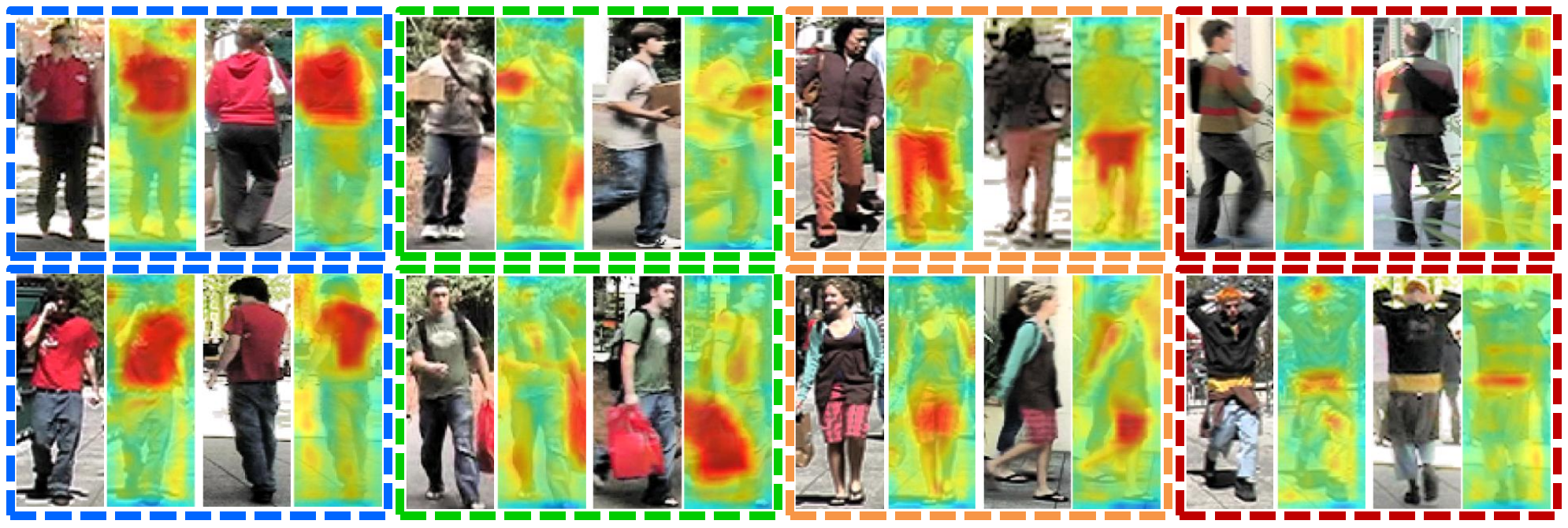} \\(a) VIPeR dataset\\\vspace{0.03in}
\includegraphics[width = 0.49\textwidth]{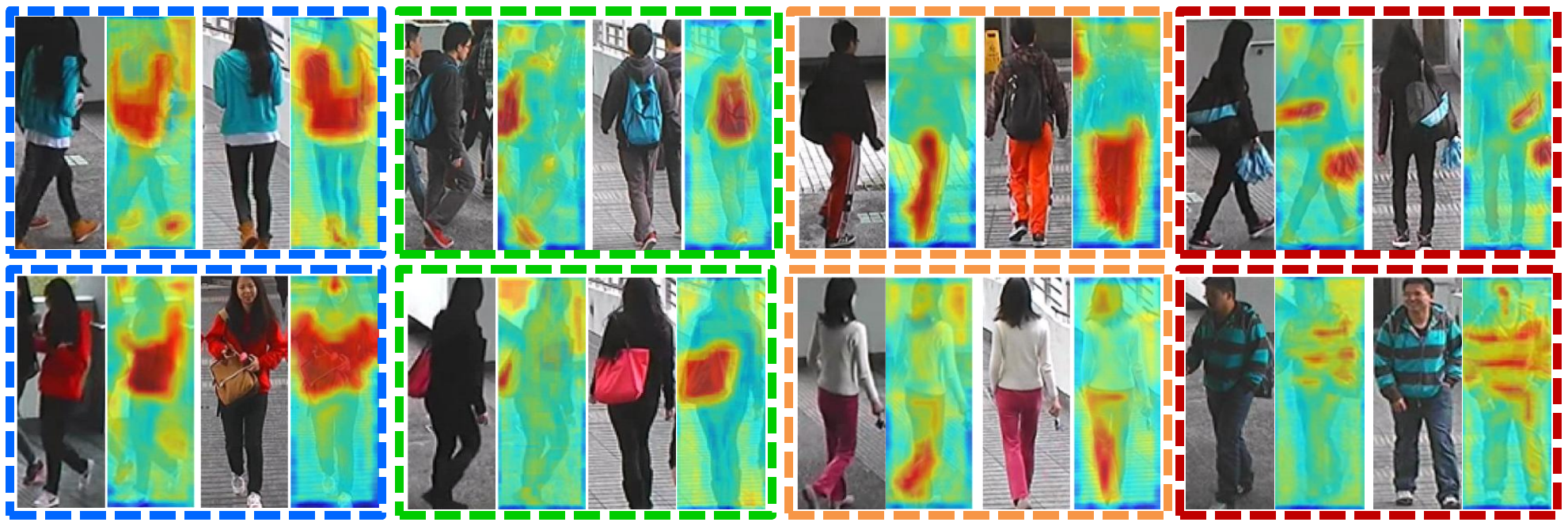} \\(b) CUHK01 dataset

\vspace{0.03in}
\caption{\small{Examples of saliency matching in our experiments. It shows four types of saliency distributions: saliency in upper body (in blue dashed box), saliency of taking bags (in green dashed box), saliency of lower body (in orange dashed box), and saliency of stripes on human body (in red dashed box). \textbf{Best viewed in color}.}}
\label{fig:salmaps}
\end{figure}

More interesting results of saliency estimation are shown in Figure \ref{fig:salmaps}(a)(b) both on the VIPeR dataset and the CUHK01 dataset. Qualitative results show our saliency learning approach could well approximate human perception and capture important salient regions on human body. 

\begin{figure}
\centering
\centering 
\includegraphics[width=0.75\linewidth]{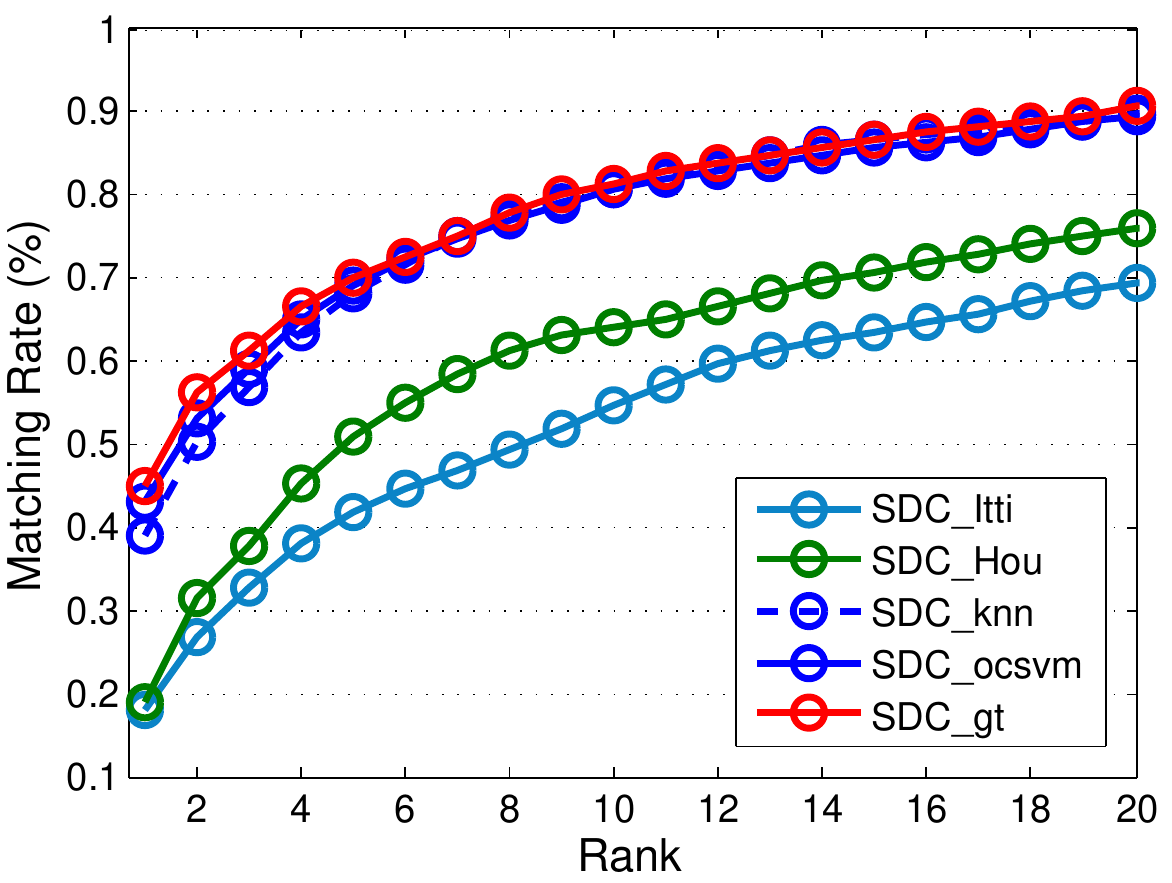} 
\caption{\small{Bi-directional weighted matching (denoted by $SDC$) using different saliency estimated by different approaches.}}

\label{fig:salvsgt}
\end{figure}

We also quantitatively compare the effectiveness of the saliency estimated from user study and our computation models in person re-identification. We regard the $100$ images (of $100$ different persons) with saliency estimated from user study as the probe set for evaluation, and images of the corresponding identities in another camera view are included as the gallery set. Bi-directional weighted matching is adopted in testing competing saliency estimation methods, including general image saliency (Itti and Hou), our learned human saliency (SDC$\_{\text{knn}}$ and SDC$\_{\text{ocsvm}}$), and saliency estimated from user study (SDC$\_{\text{gt}}$).  CMCs are reported in Figure \ref{fig:salvsgt}. Results show that the our learned human saliency can well approximate the saliency estimated from user study in person re-identification, while general image saliency significantly degrades the re-identification performance.

\subsection{Component-wise Evaluation}
\begin{figure*}[t]
\centering
\begin{minipage}{0.458\linewidth}
\centering 
\includegraphics[width = \textwidth]{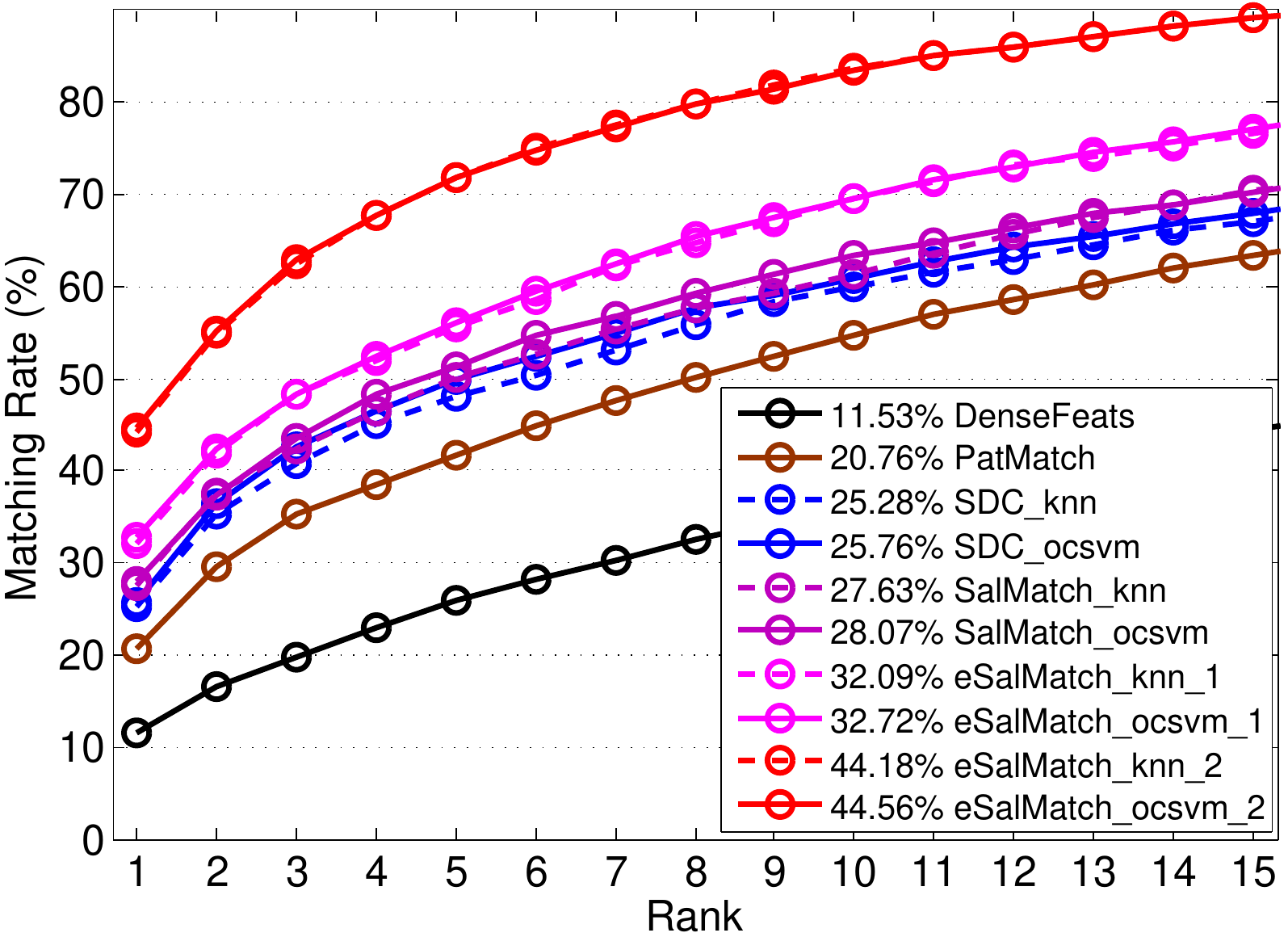}\\(a) VIPeR dataset
\end{minipage}~~~
\begin{minipage}{0.46\linewidth}
\centering 
\includegraphics[width = \textwidth]{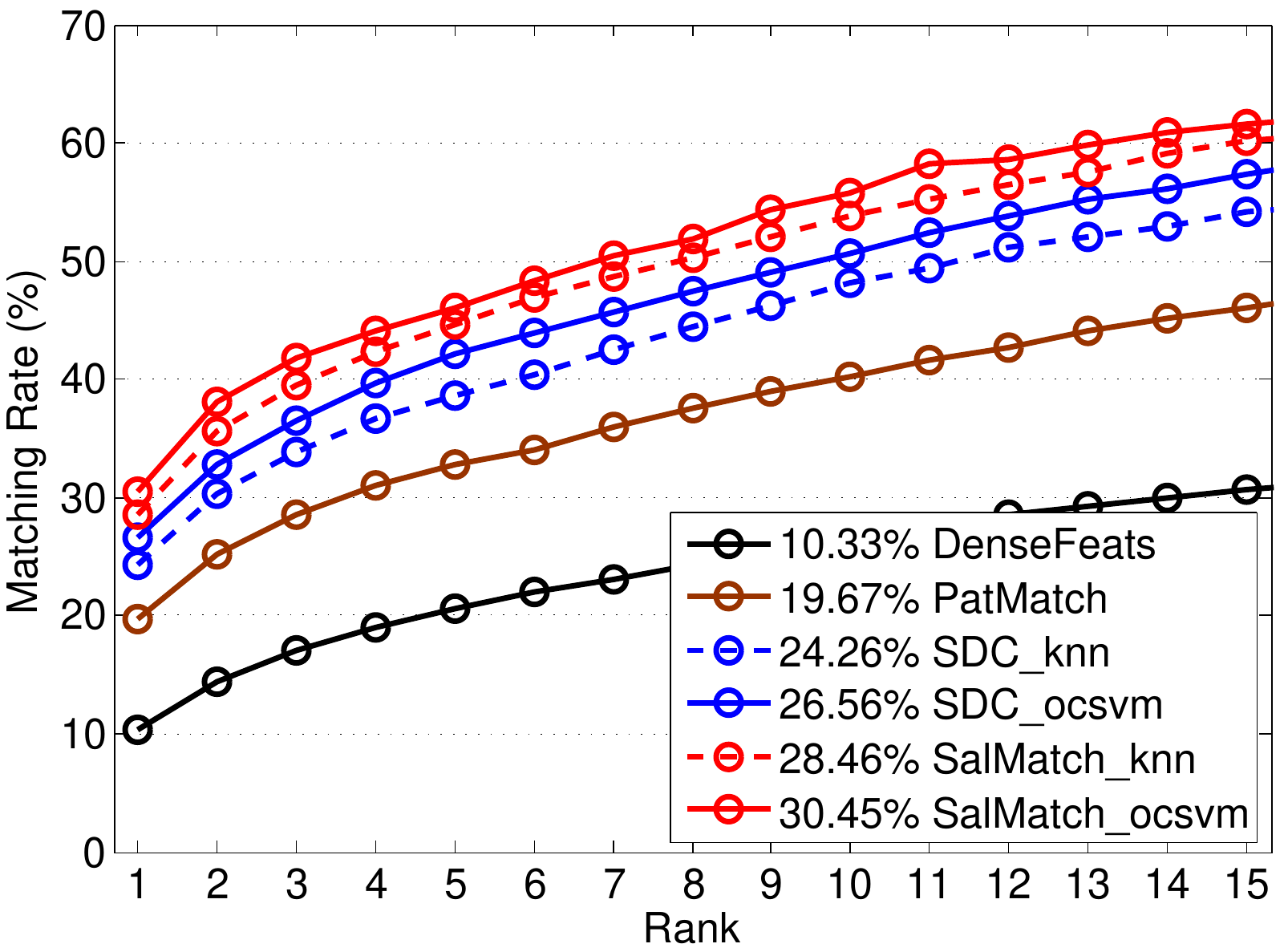} \\(b) CUHK01 dataset
\end{minipage}
    \caption{\small{CMC curves of component-wise evaluation in our approach on the VIPeR and CUHK01 datasets. 
		All the rank-1 accuracies are shown in  front of method names.}}
    \vspace{-0.1in}
    \label{fig:componentcmcs}
\end{figure*}

The effectiveness of different components in our framework is evaluated. Different settings of component combination are described in Table \ref{tab:componentsdoc} and their results are shown in Figure \ref{fig:componentcmcs}. DenseFeats in Eq. (\ref{eq:cp_densefeats}) performs the worst since it directly matches misaligned patches. PatMatch in Eq. (\ref{eq:cp_patmatch}) performs  better by handling  misalignment. SDC\_knn (SDC\_ocsvm) in Eq. (\ref{eq:cp_sdc}) improves the performance by incorporating the estimated KNN (One-class SVM) saliency in patch matching. SalMatch\_knn (SalMatch\_ocsvm) in Eq. (\ref{eq:cp_salmatch}) formulates  person re-identification as saliency matching, and learns  matching weights in a supervised way. eSalMatch\_knn\_1 (eSalMatch\_ocsvm\_1) in Eq. (\ref{eq:cp_esalmatch}) ensembles SDALF feature matching scores in SalMatch\_knn (SalMatch\_ocsvm) matching scores, and eSalMatch\_knn\_2 (eSalMatch\_ocsvm\_2) ensembles LADF similarity measures. By combining with either method, the fusion methods outperforms each component, showing that our approach is complementary to other methods. One-class SVM saliency achieves slightly better than its counterpart settings using KNN saliency.  

\small
\begin{table}
   \centering
   \resizebox{0.49\textwidth}{0.63in} {
   \begin{tabular}{r|l}
      \noalign{\hrule height 1.5pt}
      \bf Denotation       & \bf Description of component combination in test \\
      \hline
      $DenseFeats$          & Matching with concatenated patch features \\
      $PatMatch$            & Use patch matching to handle misalignment \\
      $SDC\_knn$            & Bi-directional weighted matching (KNN saliency)  \\
      $SalMatch\_knn$       & Unified saliency matching (KNN saliency) \\
      $eSalMatch\_knn\_1$      & Combine SalMatch\_knn with SDALF \cite{farenzena2010person}   \\
      $eSalMatch\_knn\_2$      & Combine SalMatch\_knn with LADF \cite{li2013learning}   \\
      $SDC\_ocsvm$          & Bi-directional weighted matching (OCSVM saliency) \\
      $SalMatch\_ocsvm$     & Unified saliency matching (OCSVM saliency) \\
      $eSalMatch\_ocsvm\_1$    & Combine SalMatch\_ocsvm with SDALF\cite{farenzena2010person} \\
      $eSalMatch\_ocsvm\_2$    & Combine SalMatch\_ocsvm with LADF\cite{li2013learning} \\
      \noalign{\hrule height 1.5pt}
   \end{tabular}
   }
   \vspace{0.1in}
   \caption{\small{Description of all the test settings in components evaluation. Refer to evaluation results in Figure \ref{fig:componentcmcs}}.}
   \label{tab:componentsdoc}
   \vspace{-0.15in}
\end{table}
\normalsize

\subsection{Comparison with the state-of-the-art}

\begin{figure*}[t]
\centering
\begin{minipage}{0.44\linewidth}
\centering 
\includegraphics[width = \textwidth]{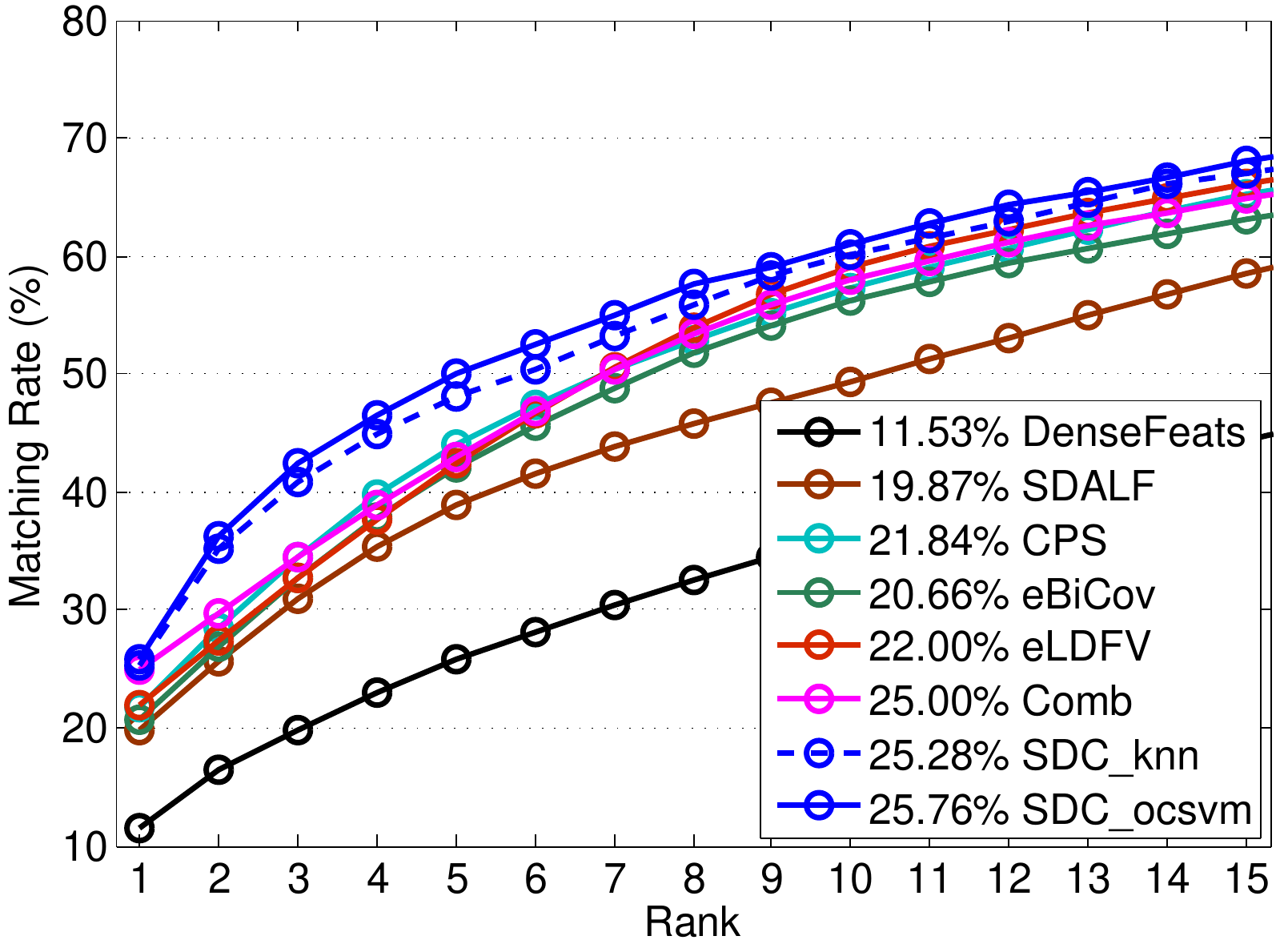}\\(a) VIPeR dataset
\end{minipage}~~~
\begin{minipage}{0.44\linewidth}
\centering 
\includegraphics[width = \textwidth]{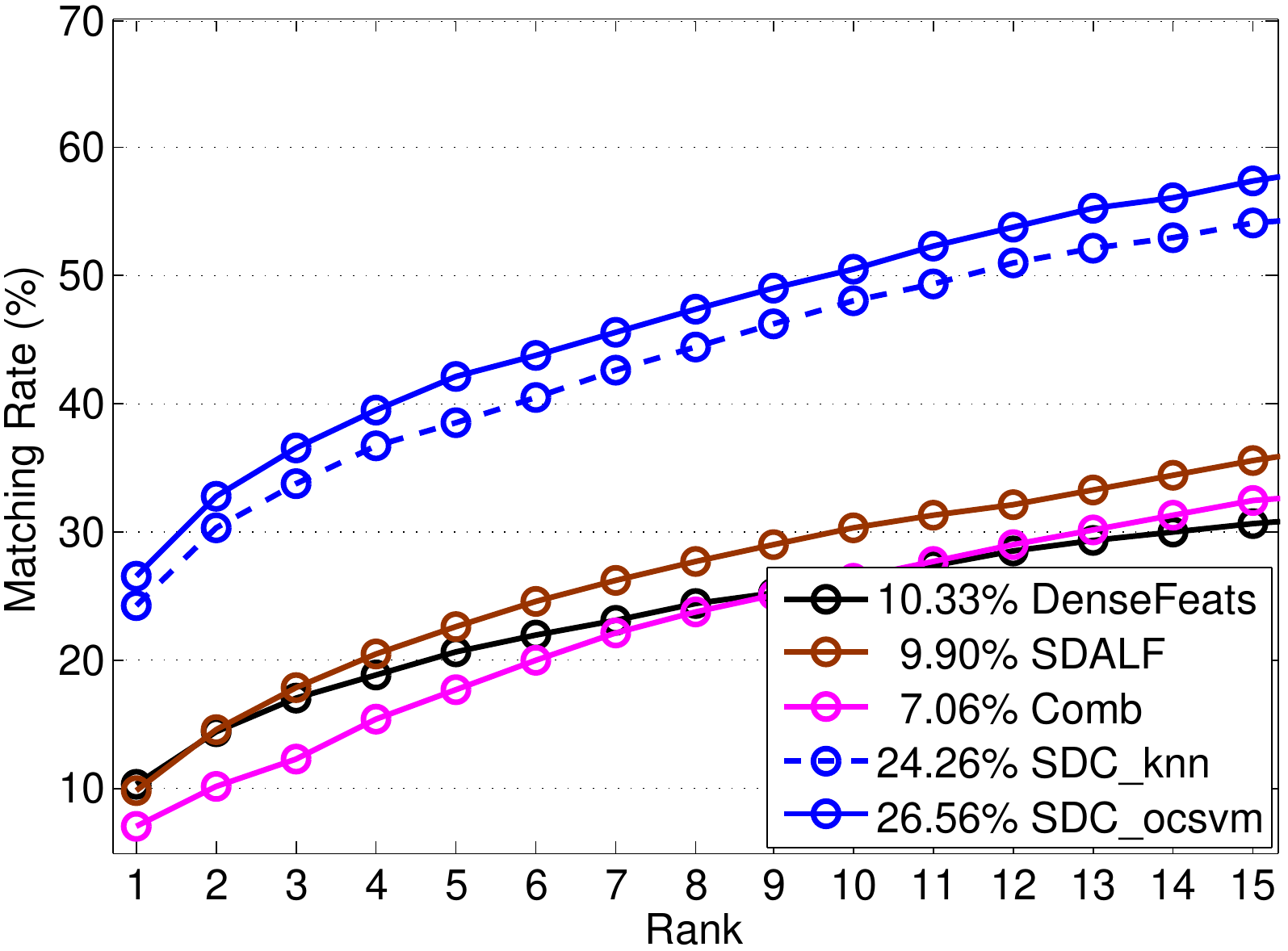} \\(b) CUHK01 dataset
\end{minipage}
    \caption{\small{CMC curves of unsupervised approaches. 
		Rank-1 accuracies are marked in front of method names.}}
    \vspace{-0.2in}
    \label{fig:unsupervised_cmcs}
\end{figure*}

Figure \ref{fig:unsupervised_cmcs} shows significant improvement of $SDC$ (unsupervised) compared with existing unsupervised methods, \emph{i.e}. SDALF \cite{farenzena2010person}, CPS \cite{cheng2011custom}, eBiCov \cite{ma2012bicov}, eLDFV \cite{malocal2012fisher}, and Comb \cite{kviatkovsky2013color} in the VIPeR dataset. For the CUHK01 dataset, we only include the $DenseFeats$ and the SDALF in comparison since code or feature representations of the other methods are not available. 

\begin{figure*}
\centering
\begin{minipage}{0.44\linewidth}
\centering 
\includegraphics[width = \textwidth]{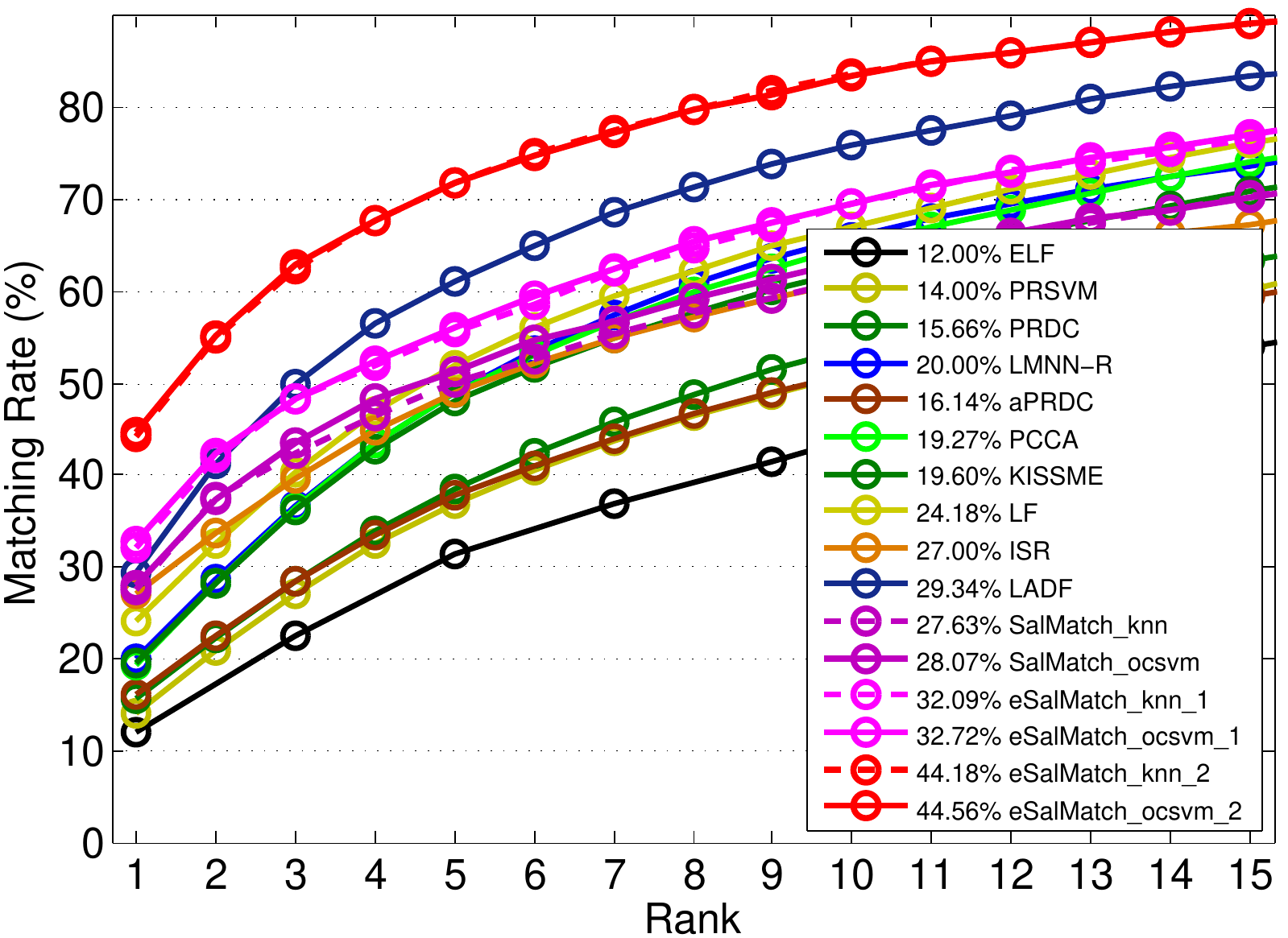}\\(a) VIPeR dataset
\end{minipage}~~~
\begin{minipage}{0.44\linewidth}
\centering 
\includegraphics[width = \textwidth]{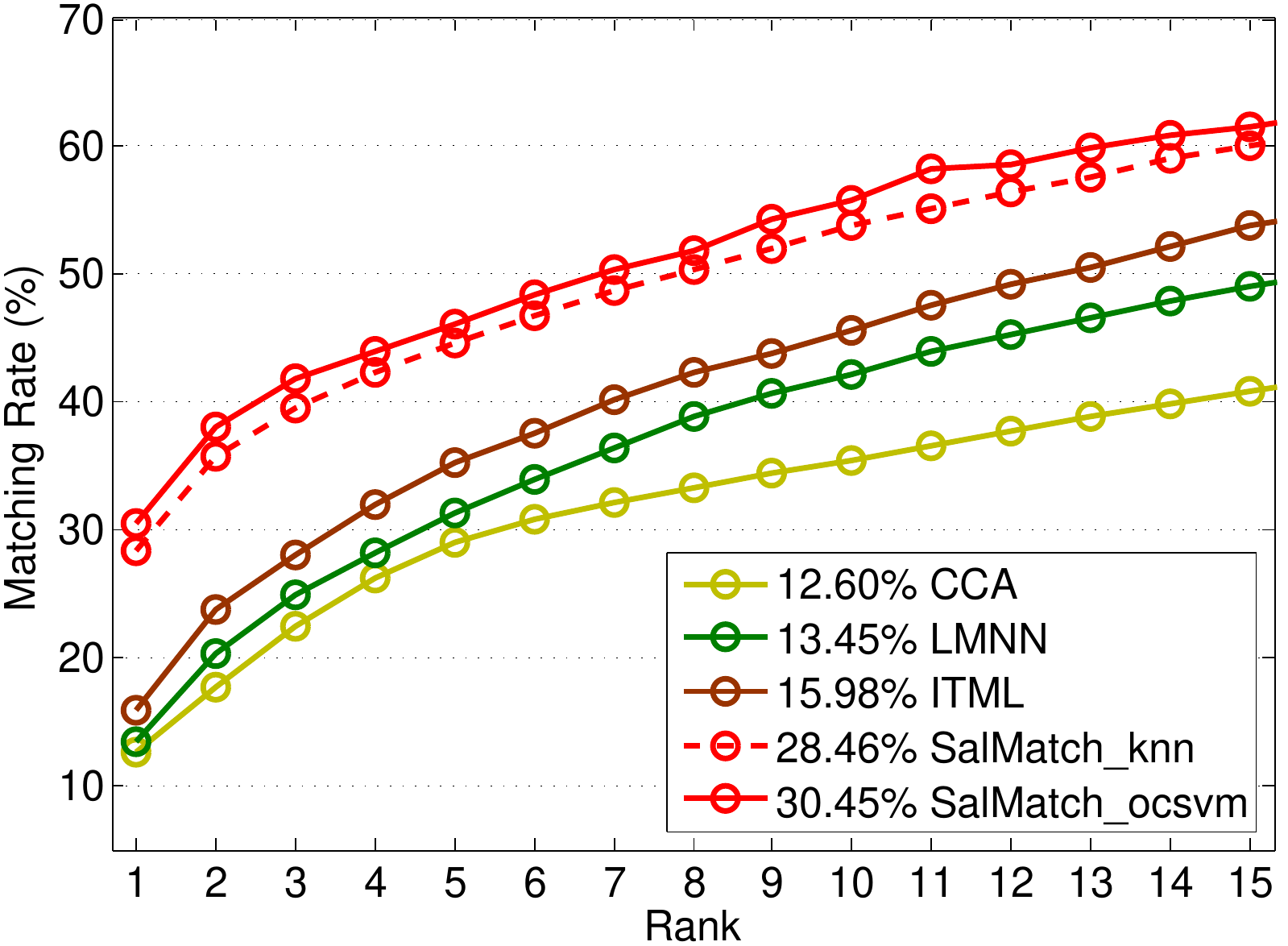} \\(b) CUHK01 dataset
\end{minipage}
    \caption{\small{CMC curves of supervised approaches.
		Rank-1 accuracies are marked in front of method names.}}
    \vspace{-0.2in}
    \label{fig:cmcs}
\end{figure*}

Figure \ref{fig:cmcs} compares our supervised saliency matching ($SalMatch$ and $eSalMatch$) with ten alternative supervised methods, including seven benchmarking distance metric learning methods, \emph{i.e}. PRDC \cite{zheng2011person}, LMNN-R \cite{dikmen2011pedestrian}, KISSME \cite{kostinger2012large}, LADF \cite{li2013learning}, PCCA \cite{mignon2012pcca}, attribute-based PRDC (aPRDC) \cite{liu2012person} and LF \cite{pedagadi2013local}, a boosting approach (ELF) \cite{gray2008viewpoint}, an ensemble of RankSVM (PRSVM) \cite{prosser2010person}, and a sparse ranking method (ISR) \cite{lisanti2014person}. Our approach outperforms all these methods. They ignore the domain knowledge on spatial variation caused by misalignment and poses as mentioned in Section \ref{sec:relatedworks}. 
Although aPRDC shares a similar spirit as ours in finding unique and inherent appearance, it weights different types global features instead of local patches. Its Rank-1 accuracy is only half of ours.
ELF has a low performance since it selects features in the original feature space in which features of different classes are highly correlated. RankSVM is similar to our method in formulating person re-identification as ranking problem. 
Combined approach $eSalMatch$ is not evaluated in CUHK01 dataset because the weights $\mu_i$ in Eq. (\ref{eq:cp_esalmatch}) are not carefully tuned for this dataset in SDALF method, and features of this dataset are not available in combining method LADF \cite{li2013learning}. 
Compared with classical metric learning methods (CCA, LMNN, and ITML) based on our $DenseFeats$ features in CUHK01 dataset, our approach also has the best performance, as shown in Figure \ref{fig:cmcs}(b).
Ours has much better performance because we adopt the discriminative saliency matching strategy for pairwise matching, and the structural SVM incorporates ranking loss in global optimization. This is implies the importance of exploiting human saliency matching and its effectiveness in training structural SVM. 



\section{Conclusion and Future Work}
We propose a novel human saliency learning and matching framework for person re-identification. Adjacency constrained patch matching is applied to build dense correspondence between image pairs to handle misalignment caused by drastic viewpoint change and pose variations. Then K-Nearest Neighbor and One-class SVM approaches are proposed to estimate saliency score for each image patch without using identity labels. User study shows that the automatically estimated human saliency has good correlation with human perception. It is more effective than general image saliency in person re-identification. The estimated saliency can be incorporated into patch matching in both the bi-directional matching scheme and the unified saliency matching framework, and images of the same identity can be recognized by maximizing the saliency matching score. Learning the weights in unified saliency matching framework is formulated as solving a structural RankSVM problem. 
Experimental results valid the effectiveness of our approach and show superior performances on both the VIPeR and CUHK01 datasets. 

The proposed framework can be extended by being integrated with other person re-identification approaches. For example, $DenseFeats$ used in this work can be replaced by other more advanced descriptors of characterizing local patches. Patch matching in our framework can be replaced by more sophisticated feature matching techniques \cite{Li_PAMI_Feature}. Since saliency information is complementary to appearance, our saliency matching result can be combined with the matching results of existing approaches to boost their performance as shown in Section \ref{ssec:combination}.    

\ifCLASSOPTIONcompsoc

\else
  \section*{Acknowledgment}
\fi




%

\bibliographystyle{IEEEtran}
\bibliography{sal}

\begin{thebibliography}{10}\itemsep=-1pt

\bibitem{achanta2012slic}
R.~Achanta, A.~Shaji, K.~Smith, A.~Lucchi, P.~Fua, and S.~Susstrunk.
\newblock Slic superpixels compared to state-of-the-art superpixel methods.
\newblock {\em IEEE Trans. on PAMI}, 34:2274--2282, 2012.

\bibitem{avraham2012learning}
T.~Avraham, I.~Gurvich, M.~Lindenbaum, and S.~Markovitch.
\newblock Learning implicit transfer for person re-identification.
\newblock In {\em Workshops ECCV}, pages 381--390. Springer, 2012.

\bibitem{avraham2014learning}
T.~Avraham and M.~Lindenbaum.
\newblock Learning appearance transfer for person re-identification.
\newblock In {\em Person Re-Identification}, pages 231--246. Springer, 2014.

\bibitem{bak2010person}
S.~Bak, E.~Corvee, F.~Br{\'e}mond, and M.~Thonnat.
\newblock Person re-identification using spatial covariance regions of human
  body parts.
\newblock In {\em Proc. AVSS}, 2010.

\bibitem{bedagkar2014survey}
A.~Bedagkar-Gala and S.~K. Shah.
\newblock A survey of approaches and trends in person re-identification.
\newblock {\em Image and Vision Computing}, 32(4):270--286, 2014.

\bibitem{borji2012exploiting}
A.~Borji and L.~Itti.
\newblock Exploiting local and global patch rarities for saliency detection.
\newblock In {\em Proc. CVPR}, 2012.

\bibitem{byers1998nearest}
S.~Byers and A.~E. Raftery.
\newblock Nearest-neighbor clutter removal for estimating features in spatial
  point processes.
\newblock {\em Journal of the American Statistical Association}, 93:577--584,
  1998.

\bibitem{carterette2006learning}
B.~Carterette and D.~Petkova.
\newblock Learning a ranking from pairwise preferences.
\newblock In {\em Proc. ACM SIGIR}, 2006.

\bibitem{chen2001one}
Y.~Chen, X.~Zhou, and T.~Huang.
\newblock One-class svm for learning in image retrieval.
\newblock In {\em Proc. ICIP}, 2001.

\bibitem{cheng2014person}
D.~S. Cheng and M.~Cristani.
\newblock Person re-identification by articulated appearance matching.
\newblock In {\em Person Re-Identification}, pages 139--160. Springer, 2014.

\bibitem{cheng2011custom}
D.~S. Cheng, M.~Cristani, M.~Stoppa, L.~Bazzani, and V.~Murino.
\newblock Custom pictorial structures for re-identification.
\newblock In {\em Proc. BMVC}, 2011.

\bibitem{dikmen2011pedestrian}
M.~Dikmen, E.~Akbas, T.~S. Huang, and N.~Ahuja.
\newblock Pedestrian recognition with a learned metric.
\newblock In {\em Proc. ACCV}, 2011.

\bibitem{farenzena2010person}
M.~Farenzena, L.~Bazzani, A.~Perina, V.~Murino, and M.~Cristani.
\newblock Person re-identification by symmetry-driven accumulation of local
  features.
\newblock In {\em Proc. CVPR}, 2010.

\bibitem{goferman2012context}
S.~Goferman, L.~Zelnik-Manor, and A.~Tal.
\newblock Context-aware saliency detection.
\newblock {\em IEEE Trans. on PAMI}, 34:1915--1926, 2012.

\bibitem{gong2013person}
S.~Gong, M.~Cristani, S.~Yan, and C.~C. Loy.
\newblock Person re-identification.
\newblock Springer, 2013.

\bibitem{gray2007evaluating}
D.~Gray, S.~Brennan, and H.~Tao.
\newblock Evaluating appearance models for recognition, reacquisition, and
  tracking.
\newblock In {\em PETS}, 2007.

\bibitem{gray2008viewpoint}
D.~Gray and H.~Tao.
\newblock Viewpoint invariant pedestrian recognition with an ensemble of
  localized features.
\newblock In {\em Proc. ECCV}, 2008.

\bibitem{heller2003one}
K.~Heller, K.~Svore, A.~Keromytis, and S.~Stolfo.
\newblock One class support vector machines for detecting anomalous windows
  registry accesses.
\newblock In {\em Workshop on Data Mining for Computer Security}, 2003.

\bibitem{hirzer2012dense}
M.~Hirzer, C.~Beleznai, M.~Kostinger, P.~M. Roth, and H.~Bischof.
\newblock Dense appearance modeling and efficient learning of camera
  transitions for person re-identification.
\newblock In {\em Proc. ICIP}, 2012.

\bibitem{hirzer2012relaxed}
M.~Hirzer, P.~M. Roth, M.~K{\"o}stinger, and H.~Bischof.
\newblock Relaxed pairwise learned metric for person re-identification.
\newblock In {\em Proc. ECCV}, 2012.

\bibitem{hou2012image}
X.~Hou, J.~Harel, and C.~Koch.
\newblock Image signature: Highlighting sparse salient regions.
\newblock {\em IEEE Transactions on Pattern Analysis and Machine Intelligence
  (PAMI)}, 34(1):194--201, 2012.

\bibitem{itti1998model}
L.~Itti, C.~Koch, E.~Niebur, et~al.
\newblock A model of saliency-based visual attention for rapid scene analysis.
\newblock {\em IEEE Trans. on PAMI}, 20:1254--1259, 1998.

\bibitem{joachims2005support}
T.~Joachims.
\newblock A support vector method for multivariate performance measures.
\newblock In {\em Proc. ICML}, 2005.

\bibitem{joachims2009cutting}
T.~Joachims, T.~Finley, and C.-N.~J. Yu.
\newblock Cutting-plane training of structural svms.
\newblock {\em Machine Learning}, 77:27--59, 2009.

\bibitem{judd2009learning}
T.~Judd, K.~Ehinger, F.~Durand, and A.~Torralba.
\newblock Learning to predict where humans look.
\newblock In {\em Proc. ICCV}, 2009.

\bibitem{kostinger2012large}
M.~Kostinger, M.~Hirzer, P.~Wohlhart, P.~M. Roth, and H.~Bischof.
\newblock Large scale metric learning from equivalence constraints.
\newblock In {\em Proc. CVPR}, pages 2288--2295. IEEE, 2012.

\bibitem{kviatkovsky2013color}
I.~Kviatkovsky, A.~Adam, and E.~Rivlin.
\newblock Color invariants for person reidentification.
\newblock {\em IEEE Trans. on PAMI}, 35(7):1622--1634, 2013.

\bibitem{layne2012towards}
R.~Layne, T.~M. Hospedales, and S.~Gong.
\newblock Towards person identification and re-identification with attributes.
\newblock In {\em Workshops ECCV}, pages 402--412. Springer, 2012.

\bibitem{layne2012person}
R.~Layne, T.~M. Hospedales, S.~Gong, et~al.
\newblock Person re-identification by attributes.
\newblock In {\em BMVC}, volume~2, page~3, 2012.

\bibitem{li2014person}
A.~Li, L.~Liu, and S.~Yan.
\newblock Person re-identification by attribute-assisted clothes appearance.
\newblock In {\em Person Re-Identification}, pages 119--138. Springer, 2014.

\bibitem{Li_PAMI_Feature}
H.~Li, X.~Huang, J.~Huang, and S.~Zhang.
\newblock Feature matching with affine-function transformation models.
\newblock {\em IEEE Transactions on Pattern Analysis and Machine Intelligence},
  36(12):2407--2422, Dec 2014.

\bibitem{li2011co}
H.~Li and K.~N. Ngan.
\newblock A co-saliency model of image pairs.
\newblock {\em IEEE Trans. on Image Processing}, 20:3365--3375, 2011.

\bibitem{liWcvpr13}
W.~Li and X.~Wang.
\newblock Locally aligned feature transforms across views.
\newblock In {\em Proc. CVPR}, 2013.

\bibitem{li2012human}
W.~Li, R.~Zhao, and X.~Wang.
\newblock Human reidentification with transferred metric learning.
\newblock In {\em Proc. ACCV}, 2012.

\bibitem{li2013learning}
Z.~Li, S.~Chang, F.~Liang, T.~S. Huang, L.~Cao, and J.~R. Smith.
\newblock Learning locally-adaptive decision functions for person verification.
\newblock In {\em CVPR}, pages 3610--3617. IEEE, 2013.

\bibitem{lisanti2014person}
G.~Lisanti, I.~Masi, A.~Bagdanov, and A.~Del~Bimbo.
\newblock Person re-identification by iterative re-weighted sparse ranking.
\newblock {\em IEEE Trans. on PAMI}, 2014.

\bibitem{liu2012person}
C.~Liu, S.~Gong, C.~C. Loy, and X.~Lin.
\newblock Person re-identification: what features are important?
\newblock In {\em Proc. ECCV}, 2012.

\bibitem{liu2013pop}
C.~Liu, C.~C. Loy, S.~Gong, and G.~Wang.
\newblock Pop: Person re-identification post-rank optimisation.
\newblock In {\em International Conference on Computer Vision}, 2013.

\bibitem{liu2012attribute}
X.~Liu, M.~Song, Q.~Zhao, D.~Tao, C.~Chen, and J.~Bu.
\newblock Attribute-restricted latent topic model for person re-identification.
\newblock {\em Pattern Recognition}, 45(12):4204--4213, 2012.

\bibitem{loy2013person}
C.~C. Loy, C.~Liu, and S.~Gong.
\newblock Person re-identification by manifold ranking.
\newblock In {\em Proc. ICIP}, volume~20, 2013.

\bibitem{ma2012bicov}
B.~Ma, Y.~Su, and F.~Jurie.
\newblock Bicov: a novel image representation for person re-identification and
  face verification.
\newblock In {\em Proc. BMVC}, 2012.

\bibitem{malocal2012fisher}
B.~Ma, Y.~Su, and F.~Jurie.
\newblock Local descriptors encoded by fisher vectors for person
  re-identification.
\newblock 2012.

\bibitem{ma2014discriminative}
B.~Ma, Y.~Su, and F.~Jurie.
\newblock Discriminative image descriptors for person re-identification.
\newblock In {\em Person Re-Identification}, pages 23--42. Springer, 2014.

\bibitem{mcfee2010metric}
B.~McFee and G.~Lanckriet.
\newblock Metric learning to rank.
\newblock In {\em Proc. ICML}, 2010.

\bibitem{mignon2012pcca}
A.~Mignon and F.~Jurie.
\newblock Pcca: A new approach for distance learning from sparse pairwise
  constraints.
\newblock In {\em Proc. CVPR}, 2012.

\bibitem{pedagadi2013local}
S.~Pedagadi, J.~Orwell, S.~Velastin, and B.~Boghossian.
\newblock Local fisher discriminant analysis for pedestrian re-identification.
\newblock In {\em Proc. CVPR}, pages 3318--3325. IEEE, 2013.

\bibitem{prosser2008multi}
B.~Prosser, S.~Gong, and T.~Xiang.
\newblock Multi-camera matching using bi-directional cumulative brightness
  transfer functions.
\newblock In {\em Proc. BMVC}, volume~8, pages 164--1. Citeseer, 2008.

\bibitem{prosser2010person}
B.~Prosser, W.-S. Zheng, S.~Gong, T.~Xiang, and Q.~Mary.
\newblock Person re-identification by support vector ranking.
\newblock In {\em Proc. BMVC}, 2010.

\bibitem{schwartz2009learning}
W.~Schwartz and L.~Davis.
\newblock Learning discriminative appearance-based models using partial least
  squares.
\newblock In {\em XXII Brazilian Symposium on Computer Graphics and Image
  Processing}, 2009.

\bibitem{vaquero2009attribute}
D.~A. Vaquero, R.~S. Feris, D.~Tran, L.~Brown, A.~Hampapur, and M.~Turk.
\newblock Attribute-based people search in surveillance environments.
\newblock In {\em WACV}, pages 1--8. IEEE, 2009.

\bibitem{vezzani2013people}
R.~Vezzani, D.~Baltieri, and R.~Cucchiara.
\newblock People reidentification in surveillance and forensics: A survey.
\newblock {\em ACM Computing Surveys (CSUR)}, 46(2):29, 2013.

\bibitem{wang2007shape}
X.~Wang, G.~Doretto, T.~Sebastian, J.~Rittscher, and P.~Tu.
\newblock Shape and appearance context modeling.
\newblock In {\em Proc. ICCV}, 2007.

\bibitem{wu2011optimizing}
Y.~Wu, M.~Mukunoki, T.~Funatomi, M.~Minoh, and S.~Lao.
\newblock Optimizing mean reciprocal rank for person re-identification.
\newblock In {\em Proc. AVSS}, pages 408--413. IEEE, 2011.

\bibitem{xu2013human}
Y.~Xu, L.~Lin, W.-S. Zheng, and X.~Liu.
\newblock Human re-identification by matching compositional template with
  cluster sampling.
\newblock In {\em Proc. ICCV}, pages 3152--3159. IEEE, 2013.

\bibitem{yue2007support}
Y.~Yue, T.~Finley, F.~Radlinski, and T.~Joachims.
\newblock A support vector method for optimizing average precision.
\newblock In {\em Proc. ACM SIGIR}, 2007.

\bibitem{zhao2014learning}
R.~Zhao, W.~Ouyang, and X.~Wang.
\newblock Learning mid-level filters for person re-identification.
\newblock In {\em Proc. CVPR}, 2014.

\bibitem{zheng2009associating}
W.-S. Zheng, S.~Gong, and T.~Xiang.
\newblock Associating groups of people.
\newblock In {\em Proc. BMVC}, 2009.

\bibitem{zheng2011person}
W.-S. Zheng, S.~Gong, and T.~Xiang.
\newblock Person re-identification by probabilistic relative distance
  comparison.
\newblock In {\em Proc. CVPR}, 2011.

\bibitem{zheng2014group}
W.-S. Zheng, S.~Gong, and T.~Xiang.
\newblock Group association: Assisting re-identification by visual context.
\newblock In {\em Person Re-Identification}, pages 183--201. Springer, 2014.

\end{thebibliography}



%
\vspace{-0.43in}
\noindent
\vspace{-0.43in}
\begin{minipage}{\linewidth}
\begin{IEEEbiography}[{\includegraphics[width=1in,height=1.2in,clip,keepaspectratio]{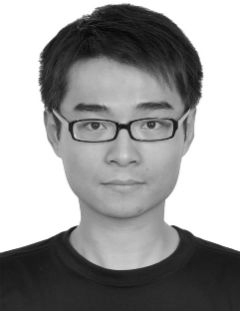}}]{Rui Zhao (S'12)}
received the B.Eng. degree in Electronic Engineering and Information Science from University of Science and Technology of China in 2010. He is currently a PhD student in the Department of Electronic Engineering at the Chinese University of Hong Kong. His research interests include computer vision, pattern recognition and machine learning. 
\end{IEEEbiography}
\end{minipage}
\vspace{-0.43in}
\begin{minipage}{\linewidth}
\begin{IEEEbiography}[{\includegraphics[width=1in,height=1.2in,clip,keepaspectratio]{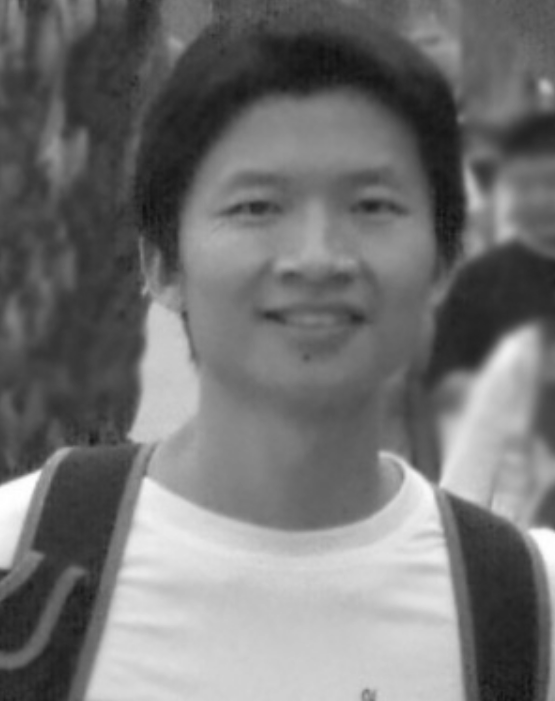}}]{Wanli Ouyang (S'08-M'11)}
received the B.S. degree in computer science from Xiangtan University, Hunan, China, in 2003. He received the M.S. degree in computer science from the College of Computer Science and Technology, Beijing University of Technology, Beijing, China. He received the PhD degree in the Department of Electronic Engineering, The Chinese University of Hong Kong, where he is now a Research Assistant Professor. His research interests include image processing, computer vision and pattern recognition. 
\end{IEEEbiography}
\end{minipage}
\vspace{-0.43in}
\begin{minipage}{\linewidth}
\begin{IEEEbiography}[{\includegraphics[width=1.2in,height=1.2in,clip,keepaspectratio]{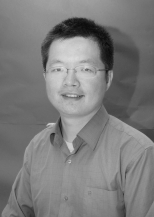}}]{Xiaogang Wang (S'03-M'10)}
received the B.S. degree from University of Science and Technology of China in 2001, the M.S. degree from Chinese University of Hong Kong in 2004, and the PhD degree in Computer Science from Massachusetts Institute of Technology. He is currently an assistant professor in the Department of Electronic Engineering at the Chinese University of Hong Kong. He received the Outstanding Young Researcher Award in Automatic Human Behaviour Analysis in 2011, and the Hong Kong Early Career Award in 2012. His research interests include computer vision and machine learning.
\end{IEEEbiography}
\end{minipage}



\end{document}